%% file: main.tex
\documentclass[10pt,twocolumn,letterpaper]{article}

\usepackage[pagenumbers]{wacv} 

\usepackage{graphicx}
\usepackage{amsmath}
\usepackage{amssymb}
\usepackage{booktabs}

\input{preamble}

\usepackage[pagebackref,breaklinks,colorlinks]{hyperref}

\usepackage[capitalize]{cleveref}
\crefname{section}{Sec.}{Secs.}
\Crefname{section}{Section}{Sections}
\Crefname{table}{Table}{Tables}
\crefname{table}{Tab.}{Tabs.}

\def\wacvPaperID{162} 
\def\confName{WACV}
\def\confYear{2025}

\begin{document}

\title{\ournetworkname: Multipurpose Video Forensics Network using Multiple Forms of Forensic Evidence}

\author{Tai D. Nguyen, Matthew C. Stamm\\
Drexel University\\
Philadelphia, PA, USA\\
{\tt\small tdn47@drexel.edu, mcs382@drexel.edu}
}
\maketitle

\input{figures/front_page_graphic_v2}

\input{sections/Abstract_v1}

\pulluppp

\input{sections/Intro_v4}

\input{sections/Background_v0}

\input{figures/overview_arch}

\input{sections/PropApproach_v2}

\input{sections/SpatialForensResidMod_v2}


\input{sections/SpatialContextMod_v1}


\input{sections/TemporalForensResidMod_v2}
\input{sections/TemporalOptFlowResidMod_v2}

\input{tables/dataset_statistics_v1}

\input{sections/MultiScaleHierTransformer_v2}


\input{sections/DetLocLoss_v0}

\input{figures/qualitative_results}

\input{sections/Data_v0}

\input{sections/Experiments_v1}

\input{sections/Discussion_v0}

\input{sections/Ablation_v0}

\input{sections/Conclusion_v0}

\subheader{Acknowledgment}
This material is based upon work supported by the National Science Foundation under Grant No. 2320600. Any opinions, findings, and conclusions or recommendations expressed in this material are those of the author(s) and do not necessarily reflect the views of the National Science Foundation.

{\small
\bibliographystyle{ieee_fullname}
\bibliography{main}
}

\end{document}

%% file: preamble.tex
\usepackage[dvipsnames]{xcolor}
\usepackage{amsmath}
\usepackage{verbatim}
\usepackage{tabularray}
\usepackage{array}
\usepackage{rotating}
\usepackage{multirow}
\usepackage{ragged2e}
\usepackage{scalerel}
\usepackage{bbm}
\usepackage{graphicx}
\usepackage{wrapfig}
\usepackage{booktabs}
\usepackage{xspace}

\newcommand{\subheader}[1]{\vspace{0.25\baselineskip} \noindent \textbf{#1.}}
\newcommand{\citethis}{\colorbox{yellow}{\makebox(22, 2.5){\color{red}\textbf{[cite]}}}}

\newcommand{\mediumcaption}{\fontsize{8.75pt}{11pt}\selectfont} 
\newcommand{\smallcaption}{\fontsize{8.5pt}{10pt}\selectfont} 

\newcommand{\smallerr}{\fontsize{6pt}{7.6pt}\selectfont}
\newcommand{\pullup}{\vspace{-0.2\baselineskip}}
\newcommand{\pullupp}{\vspace{-0.4\baselineskip}}
\newcommand{\pulluppp}{\vspace{-0.8\baselineskip}}

\newcommand{\skipless}{
	\setlength{\abovedisplayskip}{3.75pt}
	\setlength{\belowdisplayskip}{3.75pt}
	\setlength{\abovedisplayshortskip}{3.75pt}
	\setlength{\belowdisplayshortskip}{3.75pt}
}

\newcommand{\ournetworkname}{MVFNet\xspace}
\newcommand{\pooleddsname}{Unified Video Forgery Analysis\xspace}
\newcommand{\pooleddsabrv}{UVFA\xspace}

\newcommand{\best}[1]{\color{blue}{\textbf{#1}}}

\newcommand{\ldash}{\;\textemdash\,}
\newcommand{\concat}{\oplus}

\newcommand{\vframe}{I}
\newcommand{\convolve}{*}


%% file: figures/front_page_graphic_v2.tex
\begin{figure*}[!t]
	\pullupp
	\centering
	\setlength{\fboxsep}{0pt}
	\begin{minipage}[t]{1.0\textwidth}
		\setlength{\fboxrule}{0pt}
		\makebox[0.070\textwidth][s]{}
		\fbox{\parbox{0.120\textwidth}{\centering\smallerr Forged\\Video Frame}}
		\fbox{\parbox{0.120\textwidth}{\centering\smallerr Spatial Forensic\\Residuals}}
		\fbox{\parbox{0.120\textwidth}{\centering\smallerr Spatial RGB Context}}
		\fbox{\parbox{0.120\textwidth}{\centering\smallerr Temporal Forensic\\Residuals}}
		\fbox{\parbox{0.120\textwidth}{\centering\smallerr Optical Flow\\Residuals}}
		\fbox{\parbox{0.120\textwidth}{\centering\smallerr Predicted Mask}}
		\fbox{\parbox{0.120\textwidth}{\centering\smallerr Grouth-\\Truth Mask}}
		\smallskip
	\end{minipage}

	\begin{minipage}[t]{1\textwidth}
		\setlength{\fboxrule}{0pt}
		\makebox[0.070\textwidth][r]{\raisebox{14pt}{\smallerr Edit}}
		\fbox{\includegraphics[width=0.120\textwidth, height=0.068\textwidth]{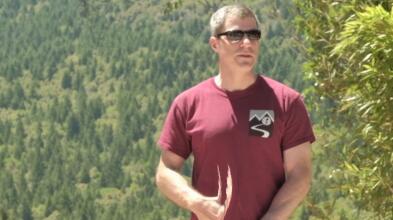}}
		\fbox{\includegraphics[width=0.120\textwidth, height=0.068\textwidth]{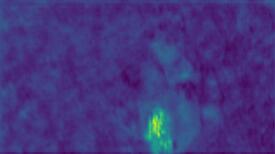}}
		\fbox{\includegraphics[width=0.120\textwidth, height=0.068\textwidth]{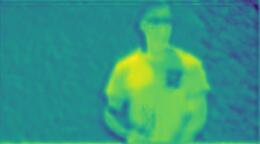}}
		\fbox{\includegraphics[width=0.120\textwidth, height=0.068\textwidth]{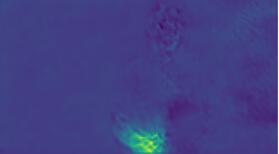}}
		\fbox{\includegraphics[width=0.120\textwidth, height=0.068\textwidth]{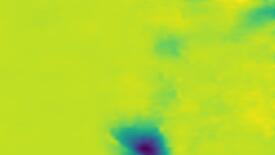}}
		\fbox{\includegraphics[width=0.120\textwidth, height=0.068\textwidth]{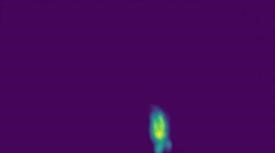}}
		\fbox{\includegraphics[width=0.120\textwidth, height=0.068\textwidth]{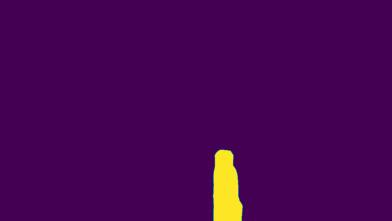}}
	\end{minipage}
	\vspace*{-0.88\baselineskip}

	\begin{minipage}[t]{1\textwidth}
		\setlength{\fboxrule}{0pt}
		\makebox[0.070\textwidth][r]{\raisebox{14pt}{\smallerr Inpaint}}
		\fbox{\includegraphics[width=0.120\textwidth, height=0.068\textwidth]{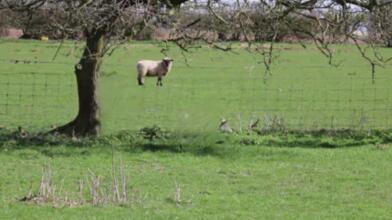}}
		\fbox{\includegraphics[width=0.120\textwidth, height=0.068\textwidth]{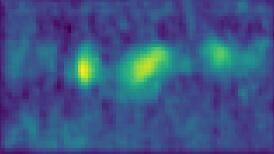}}
		\fbox{\includegraphics[width=0.120\textwidth, height=0.068\textwidth]{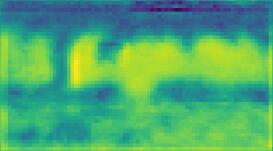}}
		\fbox{\includegraphics[width=0.120\textwidth, height=0.068\textwidth]{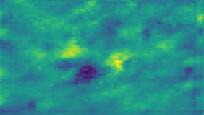}}
		\fbox{\includegraphics[width=0.120\textwidth, height=0.068\textwidth]{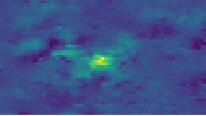}}
		\fbox{\includegraphics[width=0.120\textwidth, height=0.068\textwidth]{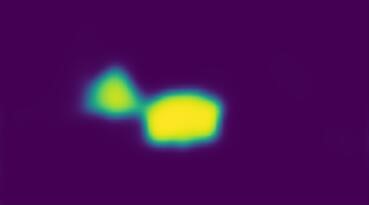}}
		\fbox{\includegraphics[width=0.120\textwidth, height=0.068\textwidth]{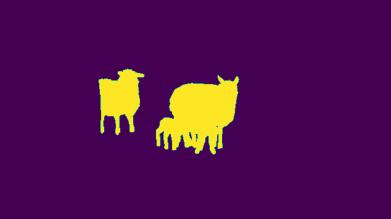}}
	\end{minipage}
	\vspace*{-0.88\baselineskip}

	\begin{minipage}[t]{1\textwidth}
		\setlength{\fboxrule}{0pt}
		\makebox[0.070\textwidth][r]{\raisebox{14pt}{\smallerr Deepfake}}
		\fbox{\includegraphics[width=0.120\textwidth, height=0.068\textwidth]{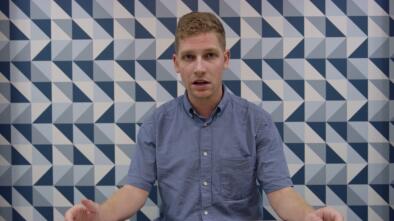}}
		\fbox{\includegraphics[width=0.120\textwidth, height=0.068\textwidth]{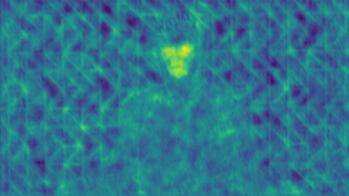}}
		\fbox{\includegraphics[width=0.120\textwidth, height=0.068\textwidth]{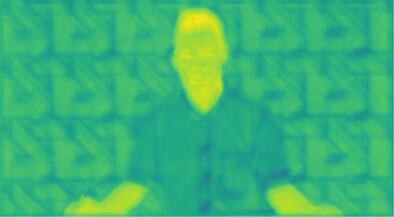}}
		\fbox{\includegraphics[width=0.120\textwidth, height=0.068\textwidth]{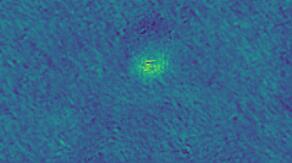}}
		\fbox{\includegraphics[width=0.120\textwidth, height=0.068\textwidth]{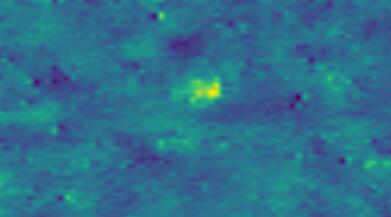}}
		\fbox{\includegraphics[width=0.120\textwidth, height=0.068\textwidth]{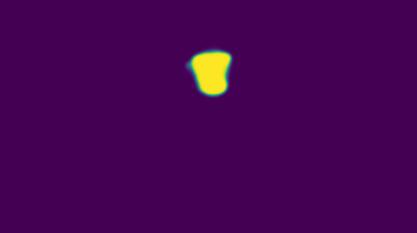}}
		\fbox{\includegraphics[width=0.120\textwidth, height=0.068\textwidth]{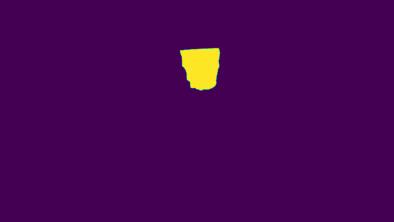}}
	\end{minipage}
	\vspace*{-1.35\baselineskip}

	\caption{
		\mediumcaption
		Examples of videos falsified using several different manipulations, alongside multiple forms of forensic evidence gathered by our network, and forgery localization masks produced by our network.
	}
	\label{fig:front_page_graphics}

	\pulluppp\pullupp
\end{figure*}

%% file: sections/Abstract_v1.tex
\begin{abstract}
	While videos can be falsified in many different ways, most existing forensic networks are specialized to detect only a single manipulation type (\eg deepfake, inpainting).  This poses a significant issue as the manipulation used to falsify a video is not known a priori.  To address this problem, we propose \ournetworkname~- a multipurpose video forensics network capable of detecting multiple types of manipulations including inpainting, deepfakes, splicing, and editing.  Our network does this by extracting and jointly analyzing a broad set
of forensic feature modalities that capture both spatial and temporal anomalies in falsified videos.
To reliably detect and localize fake content of all shapes and sizes, our network employs a novel Multi-Scale Hierarchical Transformer module to identify forensic inconsistencies across multiple spatial scales.
%
Experimental results show that our network obtains state-of-the-art performance in general scenarios where multiple different manipulations are possible, and rivals specialized detectors in targeted scenarios.


\end{abstract}

%% file: sections/Intro_v4.tex
\section{Introduction}
\label{sec:intro}

The ease with which videos can be convincingly modified or falsified poses an important misinformation threat to society. Powerful AI techniques enable the creation of visually convincing fake media.
In particular, deepfakes allow bad actors to 
spoof a person's
identity~\cite{DeepFaceLab, Deepfakes, NeuralTextures, MegaFS} and inpaiting enables the removal of 
meaningful
objects in a video~\cite{JointOpt, VINet, DFCNet, OPN}.
%
Additionally,
traditional manipulations, such as splicing and editing, can still be effective at creating plausible fake videos. For example, inauthentic content can be added through splicing and green screening, or, modified by blurring, sharpening, resizing, rotation, etc.

To combat this, researchers have created many specialized forensic authentication techniques such as: deepfake detectors~\cite{MesoNet, UCF, SBI}, inpainted video localizers~\cite{DVIL, VIDNet, YuICCV2021}, and content splicing or editing anomaly detectors~\cite{ManTra-Net, Noiseprint, MVSS-Net}.

Despite lots of research, a crucial challenge exists when creating video authentication systems in realistic scenarios: \textit{How should we detect forgeries when we don't know a priori what specific manipulation may have been used?}

While existing specialized systems work well on their targeted manipulations, they are not designed to generalize to others. 
This is because they implicitly or explicitly leverage priors about their target manipulation to aid detection.
For instance, a deepfake detector designed to identify falsified faces will fail to detect manipulations  such as green-screening or object removal using inpainting.  
Simply retraining these detectors is an imperfect, or at times infeasible solution since they are designed to capture evidence of only their target manipulation.

Therefore, it is imperative to develop video authentication techniques that can generalize over multiple forms of forgery.
To do this, we must search for evidence of forgery across a broad set of forensic feature modalities (e.g. spatially distributed forensic microstructures, temporally distributed anomalous motions),
and exploit new forensic modalities that may contain evidence of multiple manipulations.
Additionally, we cannot use any prior information about the location or content of the forgery.

In this paper, we propose \ournetworkname, a multipurpose video forensic network capable of
detecting multiple distinct manipulation types.
We do this by extracting and jointly analyzing information from multiple spatially and temporally distributed forensic feature modalities.
Important among these are temporal forensic residuals and optical flow residuals, two new forensic modalities that has not been utilized by prior work.
To reliably detect and localize fake content of all shapes and sizes, we introduce a novel multi-scale hierarchical transformer module to identify forensic inconsistencies across multiple spatial scales.
Our extensive experimental results show that \ournetworkname achieved state-of-the-art in detecting multiple different video manipulations in general scenarios and rivals specialized detectors in targeted scenarios.
This paper's novel contributions are:

\begin{itemize}
\vspace{-0.5em}
	\item We propose \ournetworkname, a new multipurpose video forensics network that is capable of detecting and localizing multiple different forms of video forgery by jointly analyzing evidence from multiple forensic modalities.  
\vspace{-0.5em}
	\item We introduce temporal forensic residuals as a new forensic feature modality that has not previously been exploited.  This modality captures temporal inconsistencies in forensic microstructures that can be introduced by several manipulations.  Additionally, we utilize optical flow residuals as an improved approach to identify motion anomalies.
\vspace{-0.5em}
	\item We create a multi-scale hierarchical transformer module that uses self-attention to identify inconsistencies in the set of joint multi-modal forensic features by leveraging information shared across multiple spatial scales.  This enables our network reliably detect and localize fake content of all shapes and sizes.
\vspace{-0.5em}
	\item We conduct a comprehensive set of experiments to show that \ournetworkname significantly outperforms existing approach's ability to detect multiple manipulations in general video forensic scenarios, as well as rivals specialized detectors on their target manipulation.
\end{itemize}
\pullupp

%% file: sections/Background_v0.tex
\pullup\pullup
\section{Background and Related Work}
\label{sec:background}
\pullup\pullup
\subheader{Video Content Forgeries}
Video manipulations have advanced from traditional splicing and editing to generative AI models like deepfakes and inpainting. Early deepfake methods~\cite{Face2Face, FaceSwap} using GANs produced inconsistent outputs, but recent approaches~\cite{FaceShifter, HifiFace, FaceDancer, SmoothSwap} now balance identity transfer and photorealism. Inpainting methods, initially object-removal tools, now employ GANs with 3D convolutions~\cite{FreeForm, LGTSM, PBVC} or attention mechanisms~\cite{STTN, CPNet, LiEECCV2020, FuseFormer}, while newer models~\cite{E2FGVI, ProPainter, FGVC} integrate optical flows for temporal consistency.

\subheader{Early Video Forensic Research}
Despite over 50 years of video development, many forensic challenges remain. Early non-deep-learning methods addressed basic issues like frame insertion/deletion, double compression, copy-move, etc., but recent studies~\cite{singh2018video, rana2022deepfake, tyagi2023detailed} show they are ineffective for modern codecs like H.264 and advanced forgeries like deepfakes or AI-guided inpainting.

\subheader{Manipulation-Specific Detection}
More recent research has focused on specialized deep-learning-based systems for specific forgeries. Deepfake detectors target artifacts like foreground versus background anomalies~\cite{WangCVPR2020, LiCVPRW2018, NiCVPR2022, ChenCVPR2022}, frequency-domain residuals~\cite{QianECCV2020, LiuCVPR2021, LuoCVPR2021}, temporal texture/identity anomalies~\cite{gu2021spatiotemporal, liu2023ti2net}, or facial motion anomalies~\cite{amerini2019deepfake, caldelli2021optical}.
Inpainting detectors focus on temporal inconsistencies~\cite{YuICCV2021, VIDNet} or unnatural motions~\cite{DVIL, DingISCAS2021}.
While effective within their domains, these methods struggle with out-of-domain forgeries~\cite{Shelke2021MTA, Nabi2022MS, Tolosana2020IF}.
This limitation is evident when adapting image-based solutions to videos, where forensic traces are different and non-uniformly distributed~\cite{VideoFACT, Jiang_2018_TIFS, Mandelli_2020_WIFS, Stamm_2010_ICIP}.
%

\subheader{Generalist Solutions}
One of a few notable
a generalist approach is VideoFACT~\cite{VideoFACT}. This method addressed the spatially uneven forensic traces problem in videos. It generates a novel type of trace, which approximates compression strength and forensic information's quality, and uses it to contextualize forensic traces to reduce false alarms and improve detection. However, VideoFACT does not handle temporal inconsistencies nor leverage multi-scale analysis.

%% file: figures/overview_arch.tex
\begin{figure*}[!t]
	\pullupp
	\centering
	\includegraphics[width=0.79\linewidth]{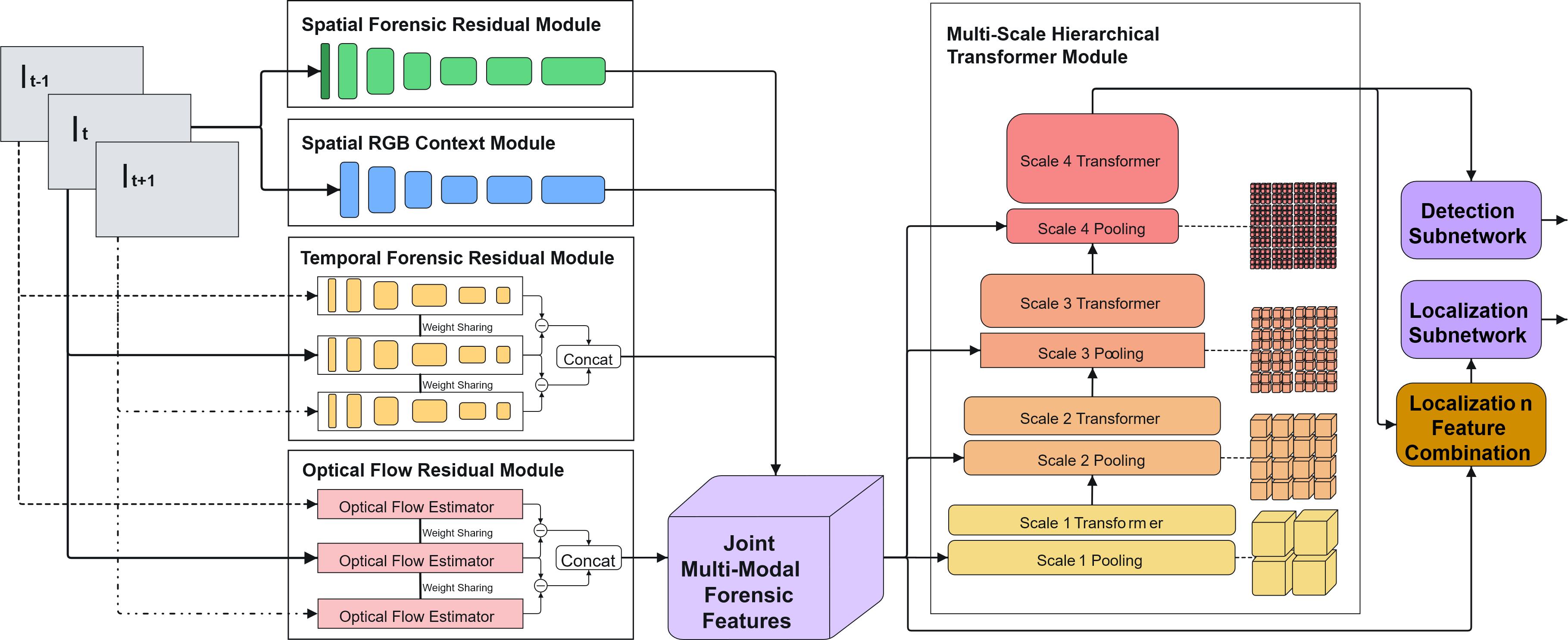}

    \pullupp

	\caption{
		\smallcaption
		Overview of \ournetworkname. Our network extracts different types (modalities) of forensic evidence: spatial forensic residuals, RGB context, and temporal forensic residuals, and optical flow residuals. Next, our network jointly analyzes all evidence using a multi-scale hierarchical transformer module. Finally, dedicated subnetworks produce final detection scores and localization masks.
	}

	\label{fig:overview_arch}

	\pulluppp\pullupp
\end{figure*}

%% file: sections/PropApproach_v2.tex
\section{Proposed Approach}
\label{sec:propApproach}

In order to detect various types of video
manipulation, it is critical to search for multiple forms of evidence that a video has been falsified.
Our network does this by using 4 specially designed forensic feature extractors to extract embedding from multiple forensic modalities: spatial forensic residuals, spatial RGB context features, temporal forensic residuals \textit{(new in forensics)}, and optical flow residuals \textit{(new in forensics)}.
%
Next, we designed a new multi-scale hierarchical transformer module to accurately detect and localize fake content regardless of their shapes and sizes.
Then, the transformer's output is then passed to two subnetworks to produce a detection score
and a localization mask.
We will describe each network component in greater detail below.

%% file: sections/SpatialForensResidMod_v2.tex
\subsection{Spatial Forensic Residual Module}
\label{subsec:spat_forens_resid_mod}

When content within a scene is edited, synthesized, or falsified, new noise-like forensic microstructures are introduced~\cite{Noiseprint, MISLnet, FSG, FSM, VideoFACT}.  
As a result, the forensic microstructures in these regions are inconsistent with those distributed throughout the rest of the video frame.
Prediction residuals are an important forensic feature modality because they suppress a frame’s content to reveal noise-like microstructures introduced by the camera and by subsequent editing.


To best capture this feature modality, 
we create a rich set of learned forensic residuals using a constrained convolutional layer proposed in~\cite{MISLnet}.
This layer is composed of a set of 5x5 kernels $\Phi^{f}$, whose output is subtracted from the center pixel value being predicted.
These learned residuals are then analyzed using a subnetwork $h_f$ consisting of a series of fused inverted residual (FIR) blocks~\cite{MobileNetV2},
to provide higher-level spatial inconsistencies information:
\skipless
\begin{equation}
	F_t= h_f(I_t\convolve(\delta-\Phi^f))
\end{equation}
where $\convolve$ denotes convolution and  $\delta$ is a unit impulse
(i.e. the convolutional identity element).
This architecture follows the ``squeeze-and-excitation'' design principle in~\cite{SqueezeAndExcitation, EfficientNetV2}, which has been shown to capture rich, class-agnostic features in earlier layers, and class-specific features in later layers.
Furthermore, to vastly improve the performance of this module, we introduce a new pre-training procedure and loss function so that the output embeddings are readily analyzed at different scales. Due to limited space, this information is presented in the supplemetary materials.

%% file: sections/SpatialContextMod_v1.tex
\subsection{Spatial RGB Context Module}
\label{subsec:spat_rgb_ctxt_mod}

It is well known that the distribution and quality of forensic residual features can change based on a number of factors, including local texture, lighting conditions, compression strength, etc.~\cite{VideoFACT, Jiang_2018_TIFS, Mandelli_2020_WIFS, Stamm_2010_ICIP}.
However, information extracted directly from  a frame's RGB values can help contextualize seemingly anomalous forensic residual features and lead the network to make more accurate decisions~\cite{VideoFACT, MVSS-Net}.  As a result, spatially distributed RGB context information forms an important forensic modality.

To exploit this information, we capture local context features that augment the forensic residual features by passing each frame 
through 
a network $h_c$ that operates directly on the frame's RGB pixel values.
This network consists of a series of FIR blocks, followed by a final $1\times1$ convolutional layer to produce the final context features $C= h_c(\vframe_t)$. 

%% file: sections/TemporalForensResidMod_v2.tex
\subsection{Temporal Forensic Residual Module}
\label{subsec:temp_forens_resid_mod}

Several types of video manipulation create fake content in each frame independently
, such as several AI-based inpainting~\cite{JointOpt, DFCNet, CPNet} or deepfake techniques~\cite{Face2Face, FaceSwap, FaceShifter, HifiFace}.
Hence, the forensic microstructures within the falsified region may not be consistent from frame-to-frame.  
This corresponds to an important form of temporally distributed evidence that, to the best of our knowledge, has not yet been explicitly exploited.
We note that the DVIL inpainting detector~\cite{DVIL}
captures temporally distributed content discrepencies as opposed to forensic microstructure changes.


To capture these temporal microstructure changes, we propose a new forensic feature modality: temporal forensic residuals.  We capture this feature modality by using a convolutional network $g_r$ to extract the microstructure information in each frame, then compute the residual difference between two frame's microstructures.  This  approximates the derivative of these microstructures with respect to time, so that temporal changes in them will be readily apparent to our network.
Specifically, we compute the forward $\overrightarrow{T_t}$ and backward $\overleftarrow{T_t}$ temporal residuals as:
\begin{equation}
	\overrightarrow{T_t} = g_r(I_t) - g_r(I_{t-1}), \;\;
	\overleftarrow{T_t} = g_r(I_t) - g_r(I_{t+1}).
\end{equation}
These are concatened into a complete set of temporal forensic residual features
$T_t = \overrightarrow{T_t} \concat \overleftarrow{T_t}$
where $\concat$ represents the concatenation operation. 

%% file: sections/TemporalOptFlowResidMod_v2.tex
\subsection{Optical Flow Residual Module}
\label{subsec:opt_flow_resid_mod}

Some video manipulations leave behind evidence in the form of small motion inconsistencies or anomalies.
For example, some deepfake systems inadvertently introduce jittery motion in certain facial features due to difficulty aligning the target and reference faces.  Similarly, many inpainting techniques cannot ensure high degrees of consistent motion within the inpainted region over long periods of time.
%
Some existing deepfake and inpainting detectors directly analyze optical flow to identify anomalous motion patterns~\cite{VIDNet, saikia2022hybrid, caldelli2021optical, chintha2020leveraging}. However, these approaches may inadvertently learn motion models specifically for content present in the training data.
As a result, they can perform poorly on out-of-distribution data. Instead, we would like our module to learn generalizable,
content independent motion anomalies, such as very minor jitters.

To better capture motion traces, we utilize optical flow residuals as one of our forensic feature modalities.
By analyzing optical flow residuals, we suppress consistent motion and amplify motion changes, including small ones that form important forensic evidence.
For a frame at time $t$, we compute the forward optical flow residual, $\overrightarrow{O_t} = \nu (I_{t-1},I_t) - \nu (I_{t-2},I_{t-1})$, and backward optical flow residual, $\overleftarrow{O_t} = \nu (I_{t+1},I_t) - \nu (I_{t+2}, I_{t+1})$,
where $\nu (X, Y)$  computes the optical flow from $X$ to $Y$. In practice, we use RAFT~\cite{RAFT}.
We then concatenate forward and backward residuals to obtain a complete set of optical flow residual features,
$O_t = \overrightarrow{O_t} \concat \overleftarrow{O_t}$.

%% file: tables/dataset_statistics_v1.tex
\definecolor{Mercury}{rgb}{0.913,0.913,0.913}
\begin{table*}[!t]
	\centering
	\caption{Statistics for in- and out-of-distribution subsets contained in the \pooleddsname (UVFA) dataset. CRF stands for Compression Rate Factor.}
	\label{table:dataset_statistics}
	\pullup
	\resizebox{0.8\linewidth}{!}{
		\begin{tblr}{
				width = \linewidth,
				colspec = {|m{9.5mm}|m{42.5mm}|m{13mm}|m{26.7mm}|m{23mm}|m{24mm}|m{7mm}|m{18mm}|},
				column{2-3} = {l},
				column{4-8} = {r},
				row{8-16} = {Mercury},
				cell{2}{1} = {r=5}{c},
				cell{7}{1} = {c=2}{0.277\linewidth},
				cell{8}{1} = {r=8}{c},
				cell{16}{1} = {c=2}{0.277\linewidth},
				vlines,
				hline{1-2,7-8,16-17} = {-}{},
			}
			& \textbf{Dataset Source} & \textbf{Manip. Type} & \textbf{Train Frames (Videos)/CRF} & \textbf{Val Frames (Videos)/CRF} & \textbf{Test Frames (Videos)/CRF} & \textbf{Train CRF} & \textbf{Test CRF}\\
			\rotatebox{90}{\parbox{20mm}{\centering \textbf{In-}\\\textbf{distribution}}}
			& VideoFACT Std. Manips.~\cite{VideoFACT} & Splicing & 180,000 (6,000) & 18,000 (600) & 36,000 (1,200) & 0, 23 & 0, 10, 23, 28\\
			& VideoFACT Std. Manips.~\cite{VideoFACT} & Editing & 180,000 (6,000) & 18,000 (600) & 36,000 (1,200) & 0, 23 & 0, 10, 23, 28\\
			& FaceForensics++~\cite{FF++} & Deepfake & 90,000 (1,538) & 9,000 (154) & 18,000 (308) & 0, 23 & 0, 10, 23, 28\\
			& DeepFakeDetection~\cite{DFD} & Deepfake & 90,000 (552) & 9,000 (52) & 18,000 (112) & 0, 23 & 0, 10, 23, 28\\
			& DEVIL Dataset~\cite{DEVIL} & Inpainting & 100,922 (1,150) & 9,620 (110) & 20,802 (240) & 0, 23 & 0, 10, 23, 28\\
			\textbf{Total per CRF} &  & \textbf{Various} & \textbf{640,922 (15,240)} & \textbf{63,620 (1,561)} & \textbf{128,802 (3060)} & 0, 23 & 0, 10, 23, 28\\
			\rotatebox{90}{\parbox{20mm}{\centering \textbf{Out-of-}\\\textbf{distribution}\\\textbf{(test-only)}}}
			& DAVIS Std. Manips. & Splicing & N/A & N/A & 14,000 (200) & N/A & 23 \\
			& DAVIS Std. Manips. & Editing & N/A & N/A & 14,000 (200) & N/A & 23 \\
			& VideoSham~\cite{VideoSham} & Various & N/A & N/A & 16,964 (175) & N/A & Unknown \\
			& FF++ FaceSwap~\cite{FF++} & Deepfake & N/A & N/A & 18,000 (308) & N/A & 23 \\
			& FF++ NeuralTexture~\cite{FF++} & Deepfake & N/A & N/A & 18,000 (308) & N/A & 23 \\
			& FaceShifter~\cite{FaceShifter} & Deepfake & N/A & N/A & 18,000 (308) & N/A & 23 \\
			& DAVIS E2FGVI-HQ~\cite{VideoFACT} & Inpainting & N/A & N/A & 6,208 (90) & N/A & 23 \\
			& DAVIS FuseFormer~\cite{VideoFACT} & Inpainting & N/A & N/A & 6,208 (90) & N/A & 23 \\
			\textbf{Total per CRF} &  & \textbf{Various} & N/A & N/A & \textbf{111,380 (1679)} & N/A & \textbf{Various}
		\end{tblr}
	}
	\pulluppp
\end{table*}

%% file: sections/MultiScaleHierTransformer_v2.tex
\subsection{Multi-Scale Hierarchical Transformer Module} 
\label{subsec:multi_scale_hier_transformer_mod}

\input{figures/multi_res_pooling}

After collecting all evidence from every forensic modality, we jointly analyze them to identify fake content.
Fake regions are typically identified by comparing forensic features aggregated over a series of analysis windows, then searching for inconsistencies.  This presents an inherent challenge: larger windows provide better estimates of forensic traces and improved false alarm suppression, at the cost of missing small or irregularly sized forgery regions.

To overcome this, we propose the novel multi-scale hierarchical transformer module shown in Fig.~\ref{fig:multi_res_pooling_arch} to analyze joint feature embeddings. This module first combines all four forensic feature modalities and pools them into embeddings at multiple spatial scales.  The set of embeddings $\psi_t^{(k)}$ at scale $k$ is formed as follows:
%
\begin{equation}
    \begin{split}
        \psi_t^{(k)}(i,j)= (F_t \concat C_t \concat T_t \concat O_t \;,\; \omega^{(k)}(i,j))
    \end{split}
\end{equation}
where $\omega^{(k)}(i,j)$ is the pooling kernel at location $(i,j)$ at scale $k$.
The size of $\omega^{(k)}$ is equal to $2^{1-k}$ times the size of the baseline kernel $\omega^{(1)}$.

This module then analyzes these embeddings in a coarse-to-fine manner.
Each scale employs a transformer, which uses its self-attention mechanism to look for meaningful disparities in the context of one another.
Furthermore, since information is shared across scales, the module can leverage the accurate localization ability of finer scales and the false alarm suppression ability of larger scales.

To do this, we define the output of the transformer at scale $k$ as $B^{(k)}$.  Its input $A^{(k)}$ is created by combining the output of the transformer at the previous scale $B^{(k-1)}$ with the pooled embeddings at the current scale $\psi_t^{(k)}$ and previous scale $\psi_t^{(k-1)}$.
Because these do not all have the same spatial dimensions, we join them
by doubling the spatial size of the embeddings at the previous scale and adding them to the embedding of the current scale:
\begin{equation}
        B_t^{(k)}= \rho(\psi_t^{(k)} + A_t^{(k)}) \convolve \kappa + \psi_t^{(k+1)}
\end{equation}
where $\rho(\cdot)$ is the zero interleaving operation and $\kappa$ is the bilinear interpolation kernel.
Our complete module is formed by stacking four scales of pooled embeddings and transformers on top of one another.

We note that while some prior networks use some form of multi-scale analysis~\cite{MVSS-Net, ManTra-Net, wang2022m2tr, liu2018image}, these approaches are noticeably less sophisticated than ours.  They do not exploit cross-scale information sharing or the self-attention mechanism provided by transformers.  Furthermore, these approaches all process information from fine-to-coarse scales.
As our ablation study shows, this is highly suboptimal as it predisposes the network to false alarms at the finest scale, which can then propagate throughout the system.


%% file: figures/multi_res_pooling.tex
\begin{figure}[!t]
	\centering
	\includegraphics[width=0.85\linewidth]{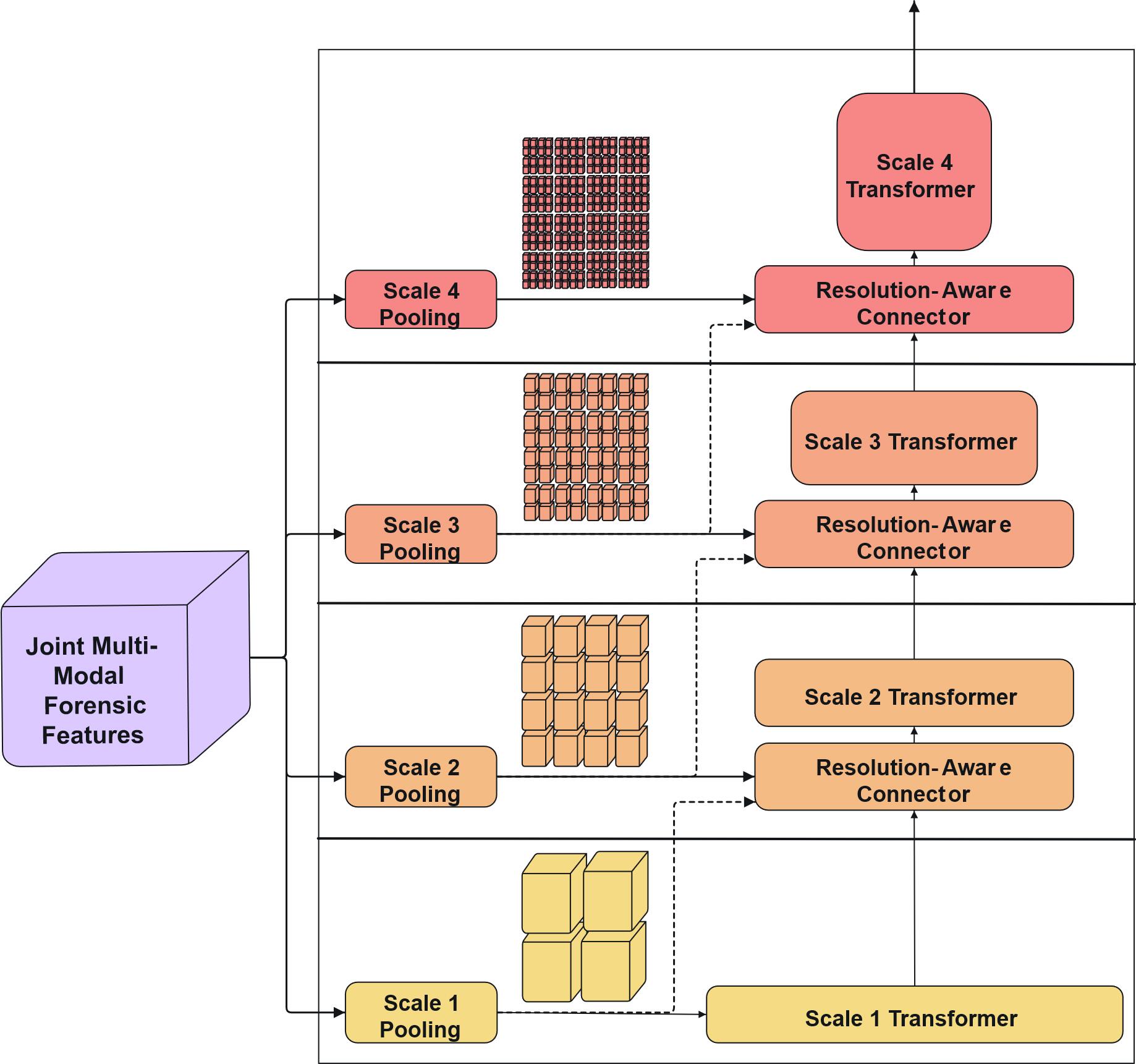}

	\caption{
		\smallcaption
Overview of our multi-scale hierarchical transformer module.  This is made using a series of adaptive poolings, resolution-aware connectors, and transformers at multiple scales. This module is designed so that information  flows in a coarse-to-fine manner.
	}

	\label{fig:multi_res_pooling_arch}

	\pulluppp\pullupp\pullup
\end{figure}

%% file: sections/DetLocLoss_v0.tex
\subsection{Detection and Localization Modules}
\label{subsec:det_loc_mod}

%

Our network outputs: a frame-level detection score that indicates if fake content is present
and a pixel-level localization mask that specifies its location.

\subheader{Detection Subnetwork}
A detection score is obtained by passing the output of the transformer module through a series of convolutional and pooling layers, then, through a final feedforward network.

\subheader{Localization Subnetwork}
To generate a high-fidelity localization mask, we concatenate an upscaled version of the transformer output with the pre-pooling multi-modal forensic features. This is then passed through a series of convolutional layers, resulting in a sigmoided 2D map.

\subheader{Joint Loss}
We introduce a composite loss function that consists of three terms: binary cross-entropy (BCE) for frame-level detection loss, BCE for pixel-level localization loss, and Dice loss, which controls for the class imbalance between manipulated and authentic pixels
such that
\skipless
\begingroup\makeatletter\def\f@size{9.5}\check@mathfonts
\begin{align}
	{}&\mathcal{L} =
	\gamma
		\biggl[
			\frac{-1}{N} \sum_{c=1}^{N} y_c \log(p_c) + (1 - y_c) \log(1 - p_c)
		\biggl] \nonumber\\
	{}&+ \alpha
		\biggl[
			\frac{-1}{MN} \sum_{i,j=1}^{M,N} \log(\hat{m}_{ij}) + (1 - m_{ij}) \log(1 - \hat{m}_{ij})
		\biggl] \nonumber\\
	{}&+ \beta
		\biggl(
			1 - \sum_{i,j=1}^{M,N} \frac{2 \cdot m_{ij} \hat{m}_{ij}}{m_{ij}^2 + \hat{m}_{ij}^2}
		\biggl)
\end{align}\endgroup
where
$y_c$ is the frame-level label, $p_c$ is the predicted probability whether the input is real or falsified, $m$ is the ground-truth mask \& $\hat{m}$ is the predicted mask.

%% file: figures/qualitative_results.tex
\begin{figure*}[!t]

    \centering
    \setlength{\fboxsep}{0pt}
    \begin{minipage}[t]{1\textwidth}
    	\setlength{\fboxrule}{-2pt}
        \makebox[0.070\textwidth][s]{    }
        \makebox[0.0974\textwidth]{\raisebox{5pt}{\smallerr Frame}}
        \makebox[0.0974\textwidth]{\raisebox{5pt}{\smallerr GT Mask}}
        \makebox[0.0974\textwidth]{\raisebox{5pt}{\smallerr Ours}}
        \makebox[0.0974\textwidth]{\raisebox{5pt}{\smallerr VideoFACT}}
        \makebox[0.0974\textwidth]{\raisebox{5pt}{\smallerr VIDNet}}
        \makebox[0.0974\textwidth]{\raisebox{5pt}{\smallerr DVIL}}
        \makebox[0.0974\textwidth]{\raisebox{5pt}{\smallerr MVSSNet}}
        \makebox[0.0974\textwidth]{\raisebox{5pt}{\smallerr Mantra-Net}}
        \makebox[0.0974\textwidth]{\raisebox{5pt}{\smallerr FSG}}
    \end{minipage}

	\vspace*{-0.1\baselineskip}

    \begin{minipage}[t]{1\textwidth}
    	\setlength{\fboxrule}{-2pt}
        \makebox[0.070\textwidth][r]{\raisebox{8pt}{\smallerr Deepfake\hspace{2pt}}}
        \fbox{\includegraphics[width=0.1070\textwidth]{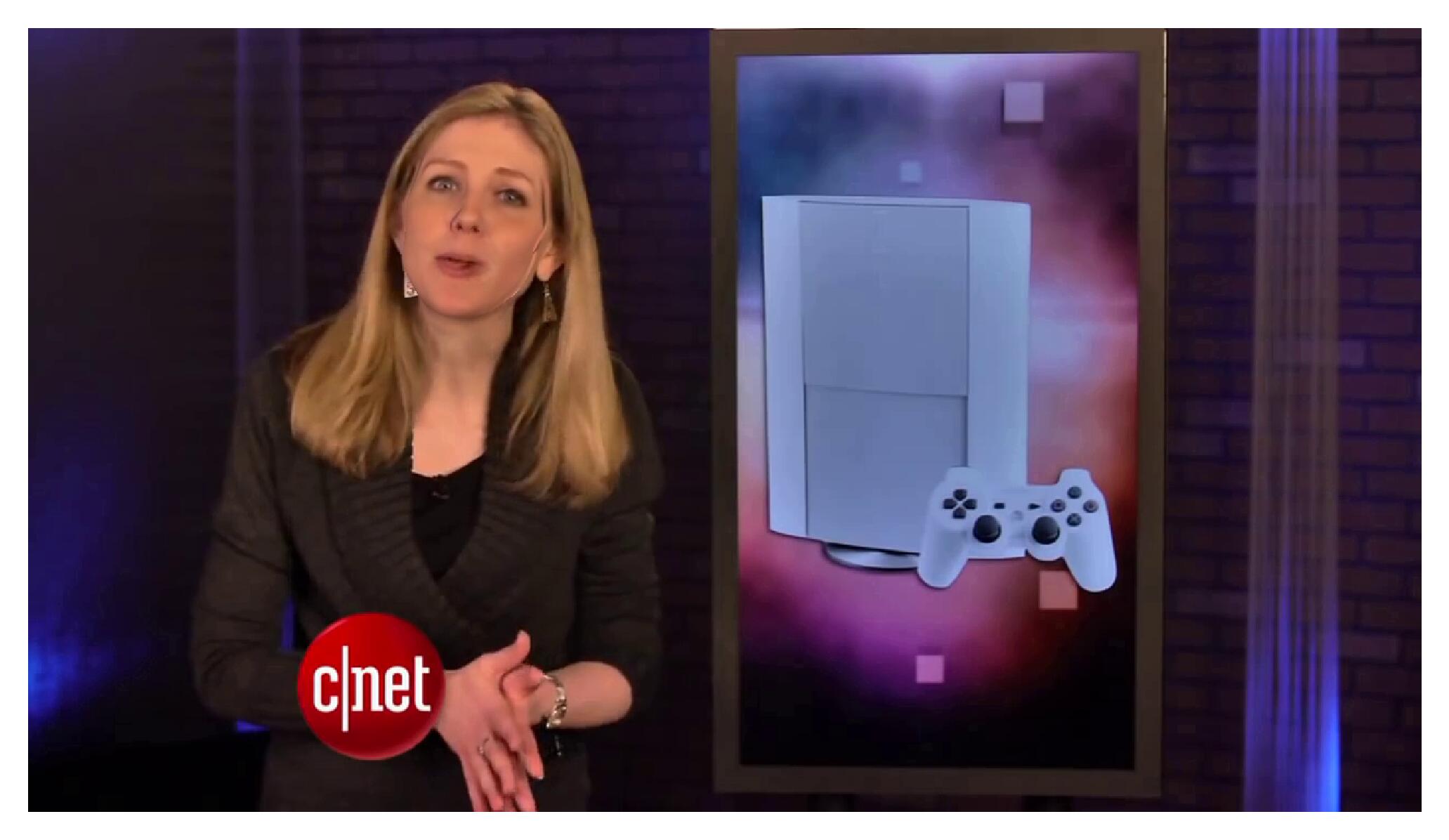}}
        \fbox{\includegraphics[width=0.1070\textwidth]{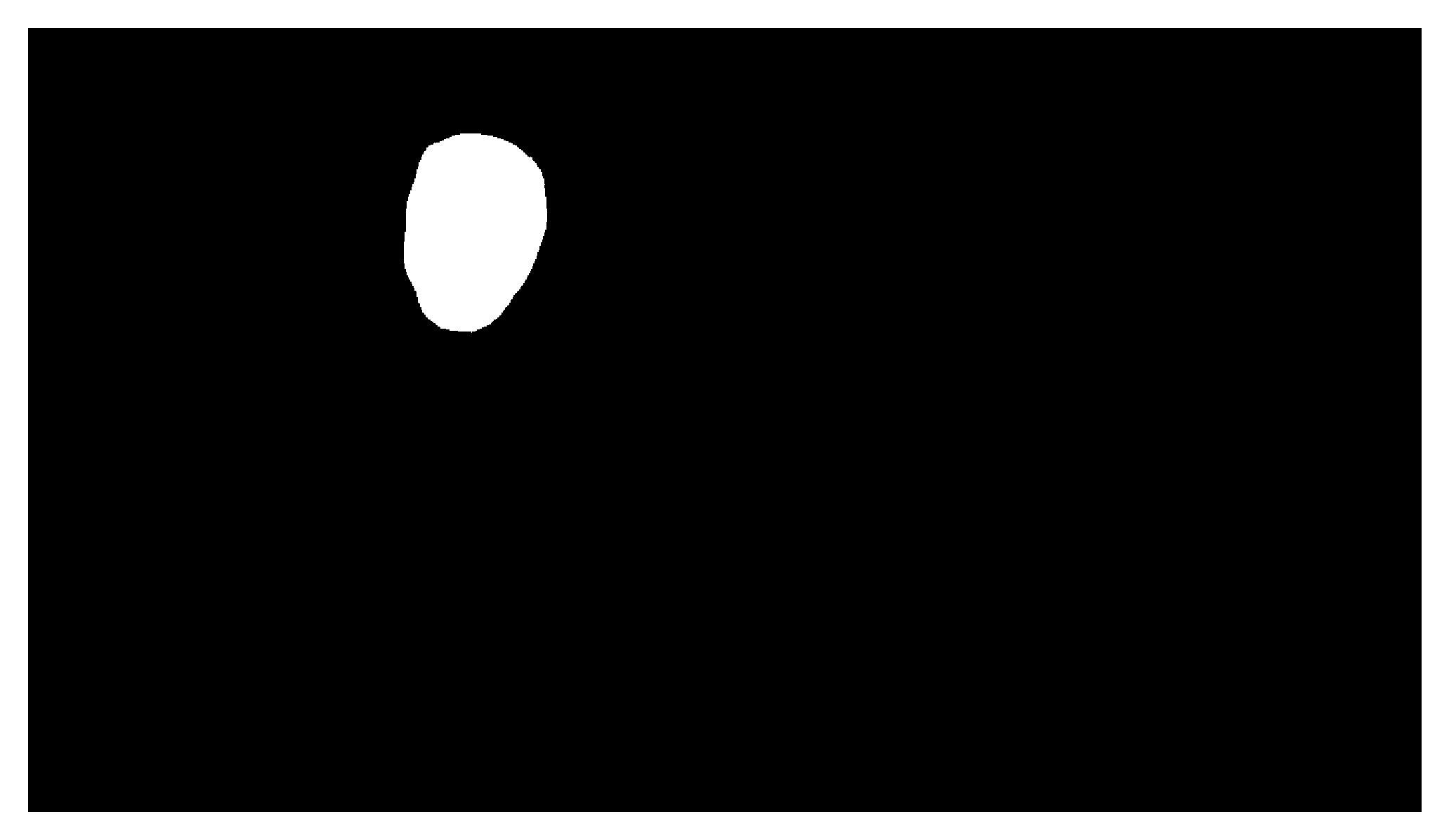}}
        \fbox{\includegraphics[width=0.1070\textwidth]{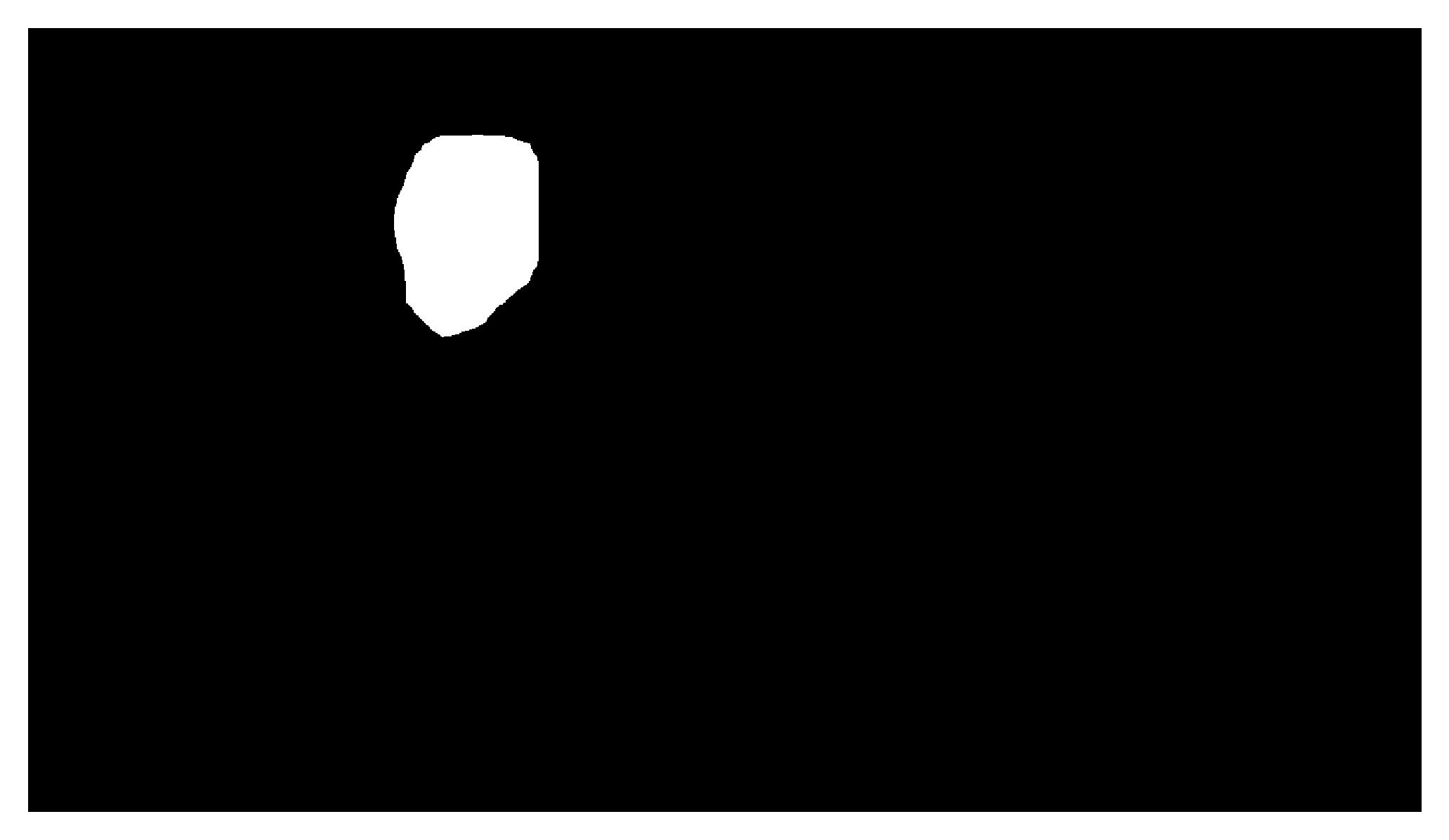}}
        \fbox{\includegraphics[width=0.1070\textwidth]{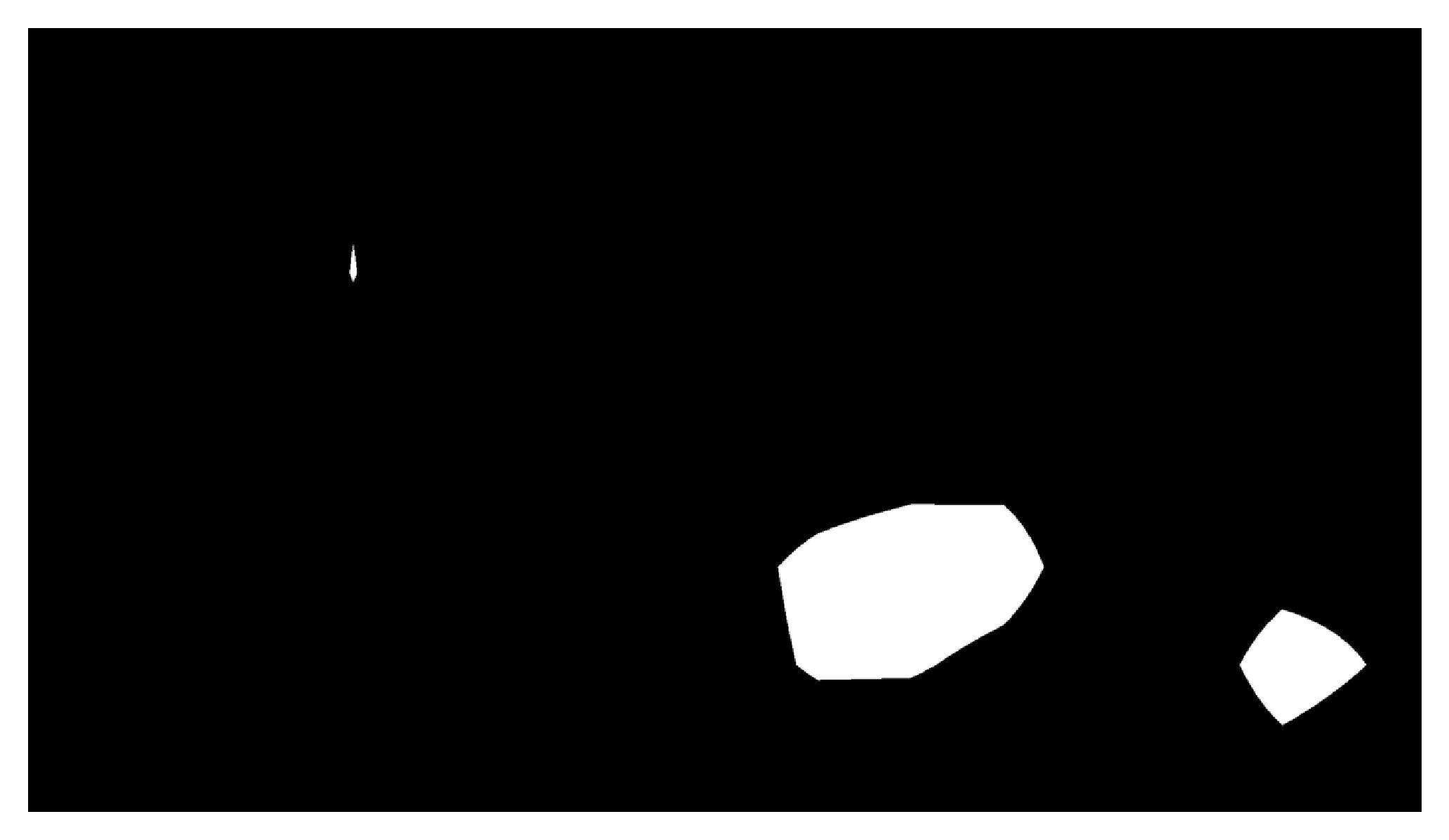}}
        \fbox{\includegraphics[width=0.1070\textwidth]{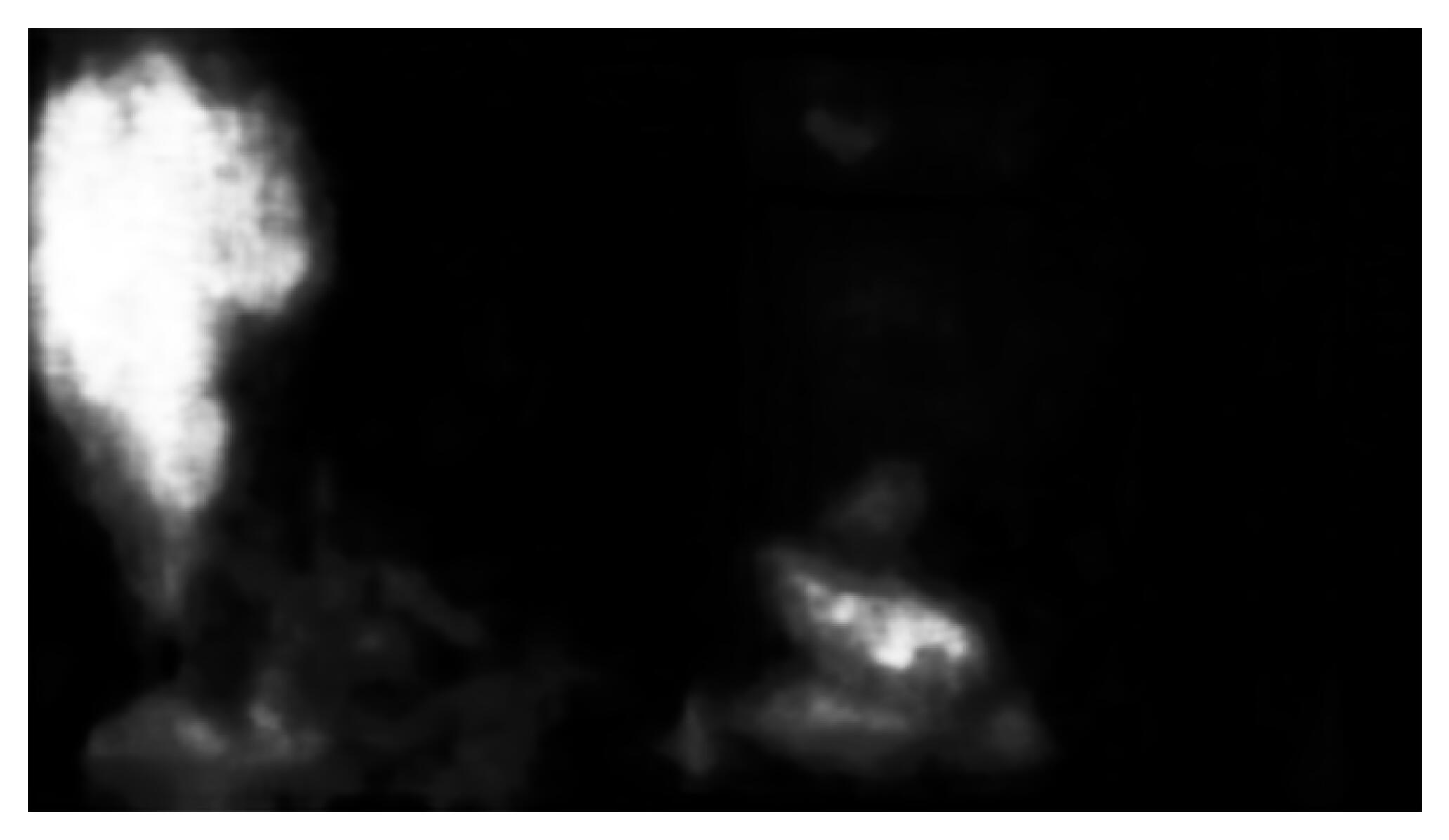}}
        \fbox{\includegraphics[width=0.1070\textwidth]{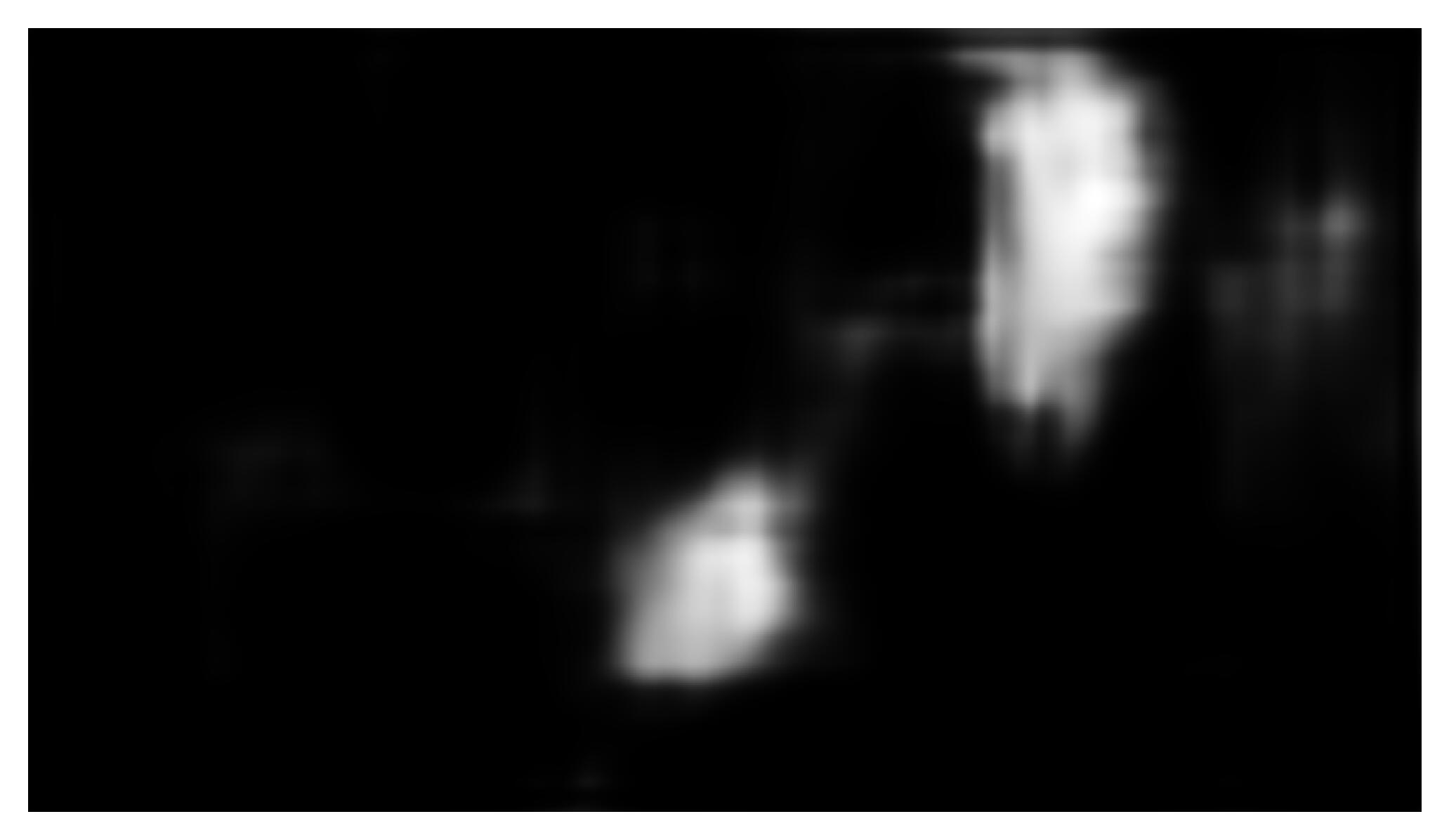}}
        \fbox{\includegraphics[width=0.1070\textwidth]{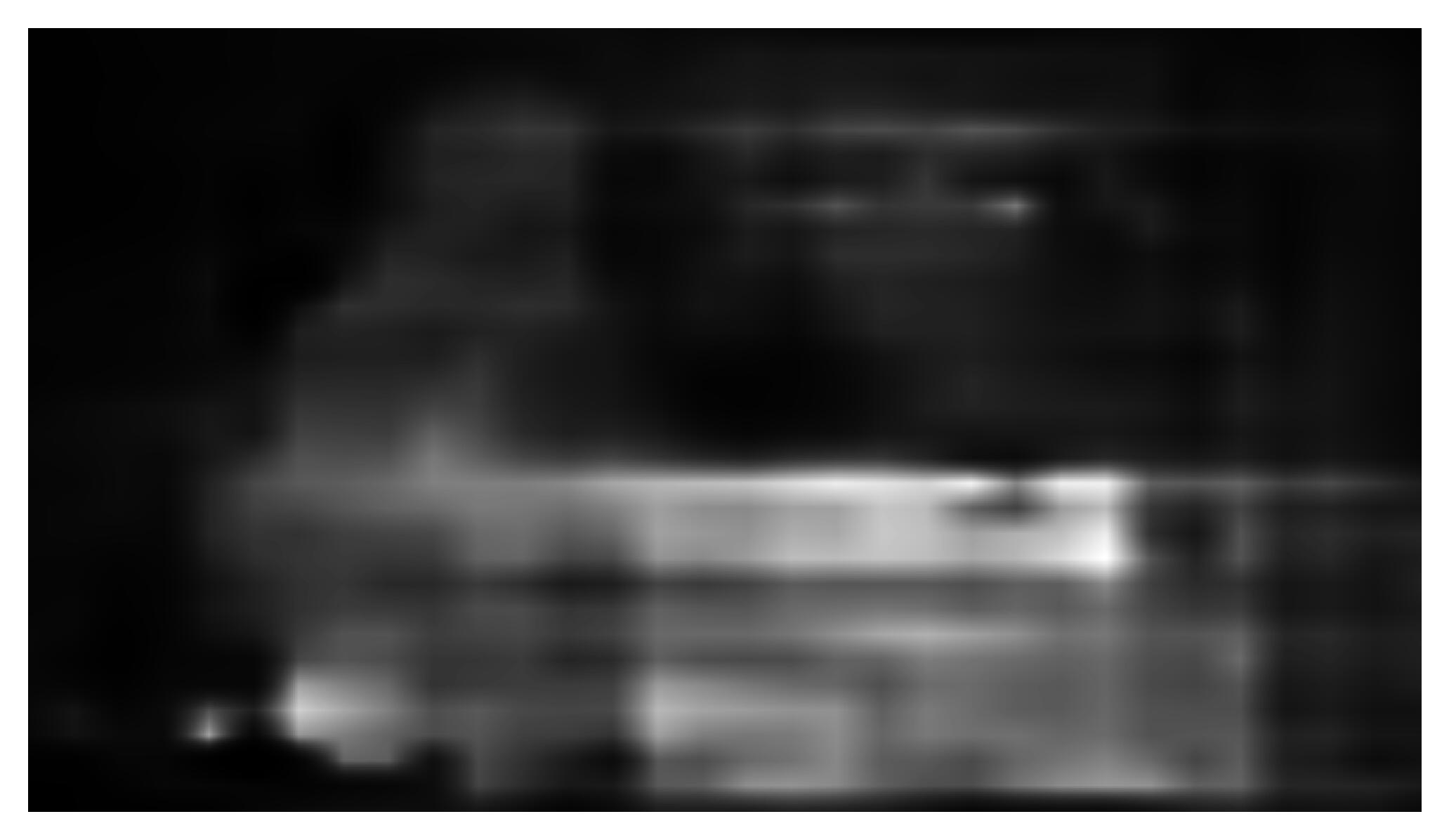}}
        \fbox{\includegraphics[width=0.1070\textwidth]{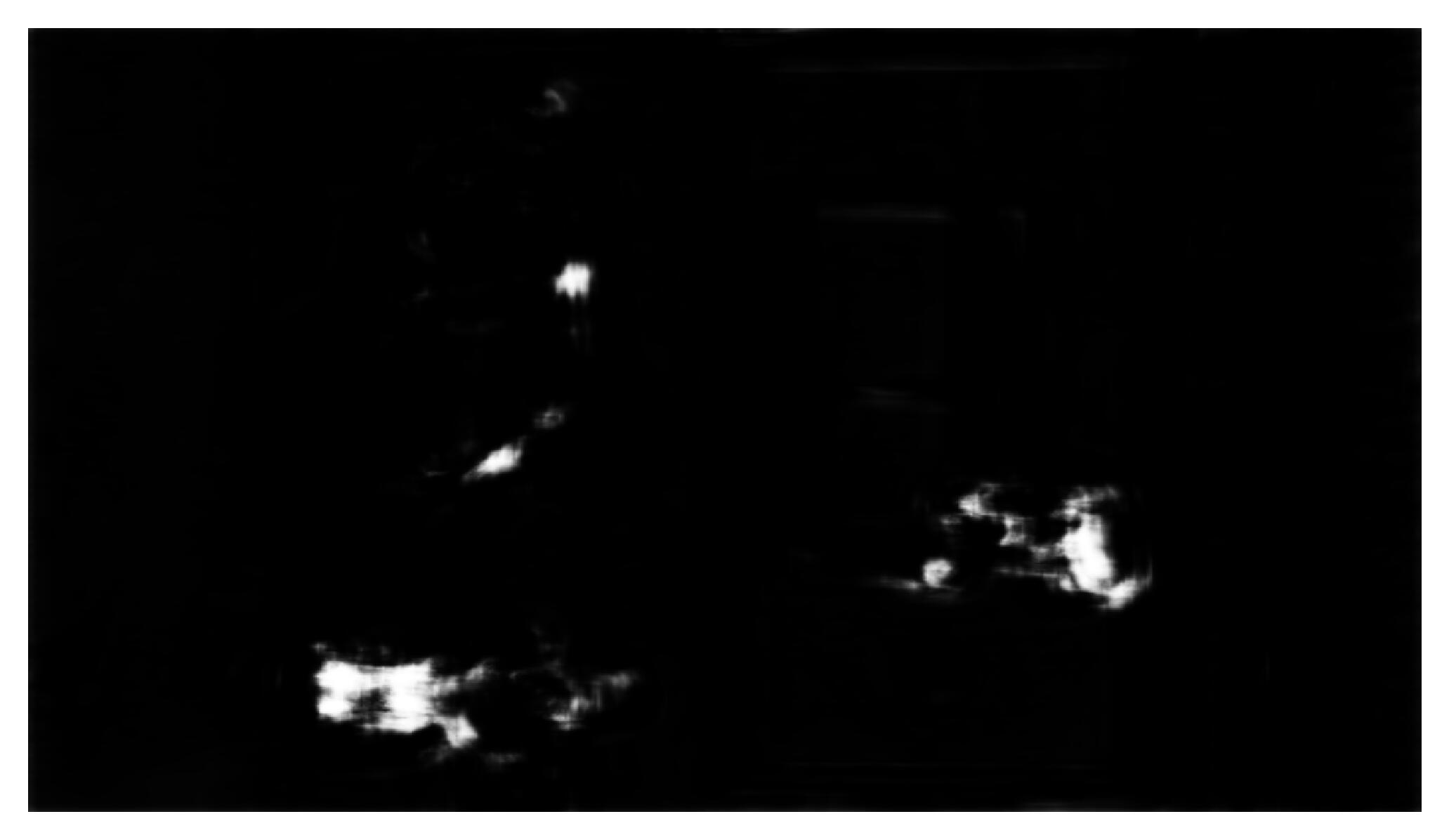}}
        \fbox{\includegraphics[width=0.1070\textwidth]{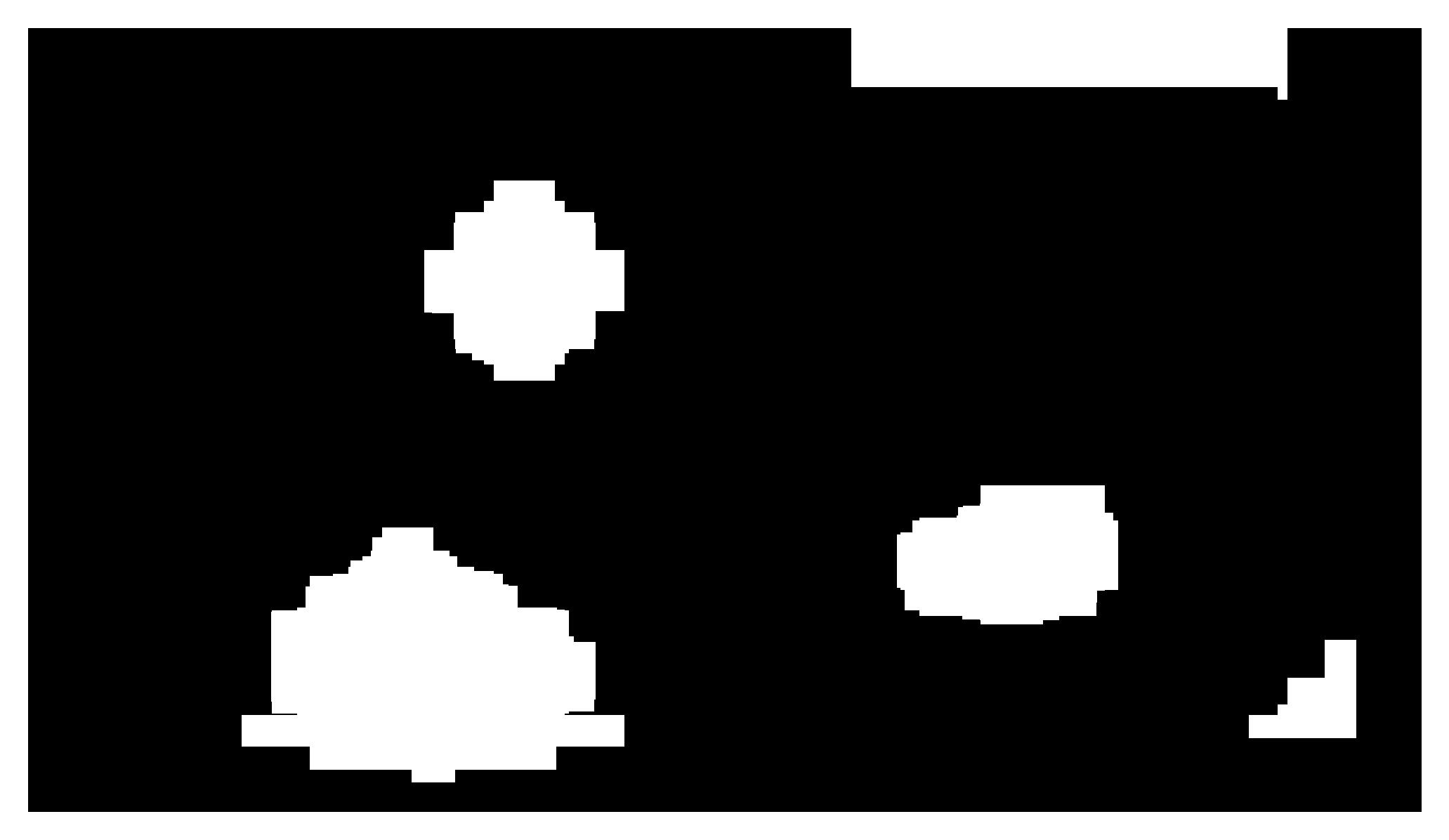}}
    \end{minipage}

	\vspace*{-0.1\baselineskip}

    \begin{minipage}[t]{1\textwidth}
    	\setlength{\fboxrule}{-2pt}
        \makebox[0.070\textwidth][r]{\raisebox{8pt}{\smallerr Inpainting\hspace{2pt}}}
        \fbox{\includegraphics[width=0.1070\textwidth]{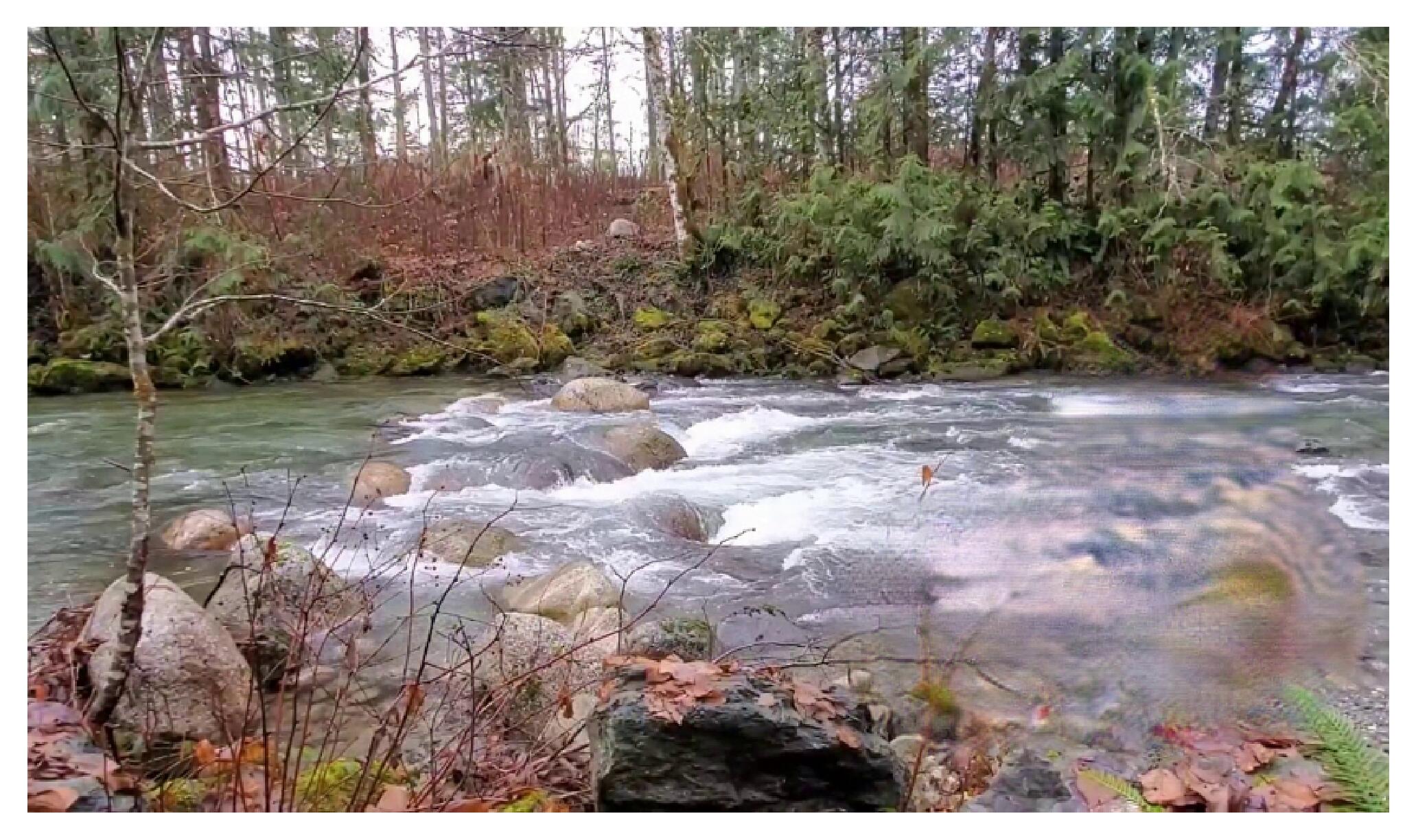}}
        \fbox{\includegraphics[width=0.1070\textwidth]{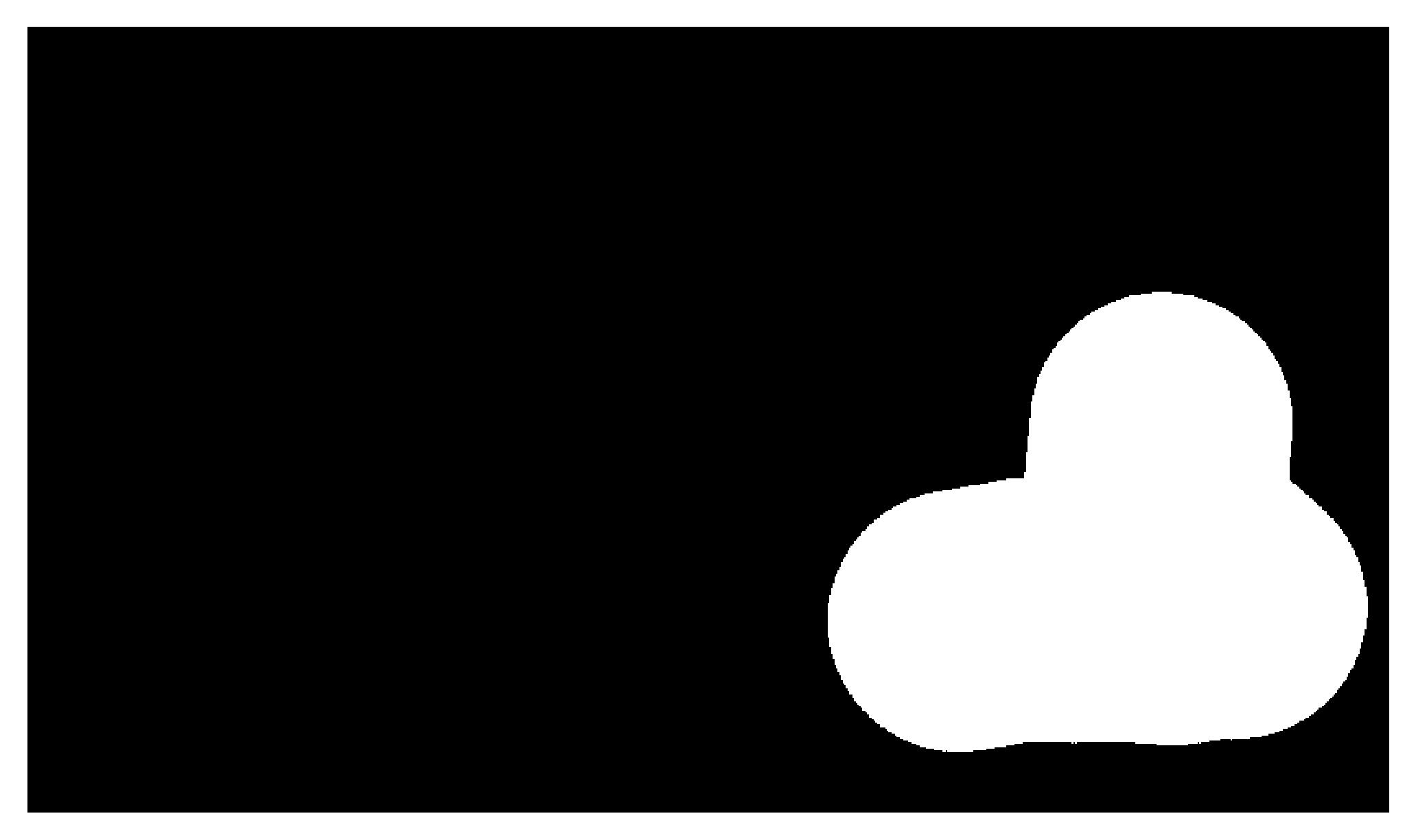}}
        \fbox{\includegraphics[width=0.1070\textwidth]{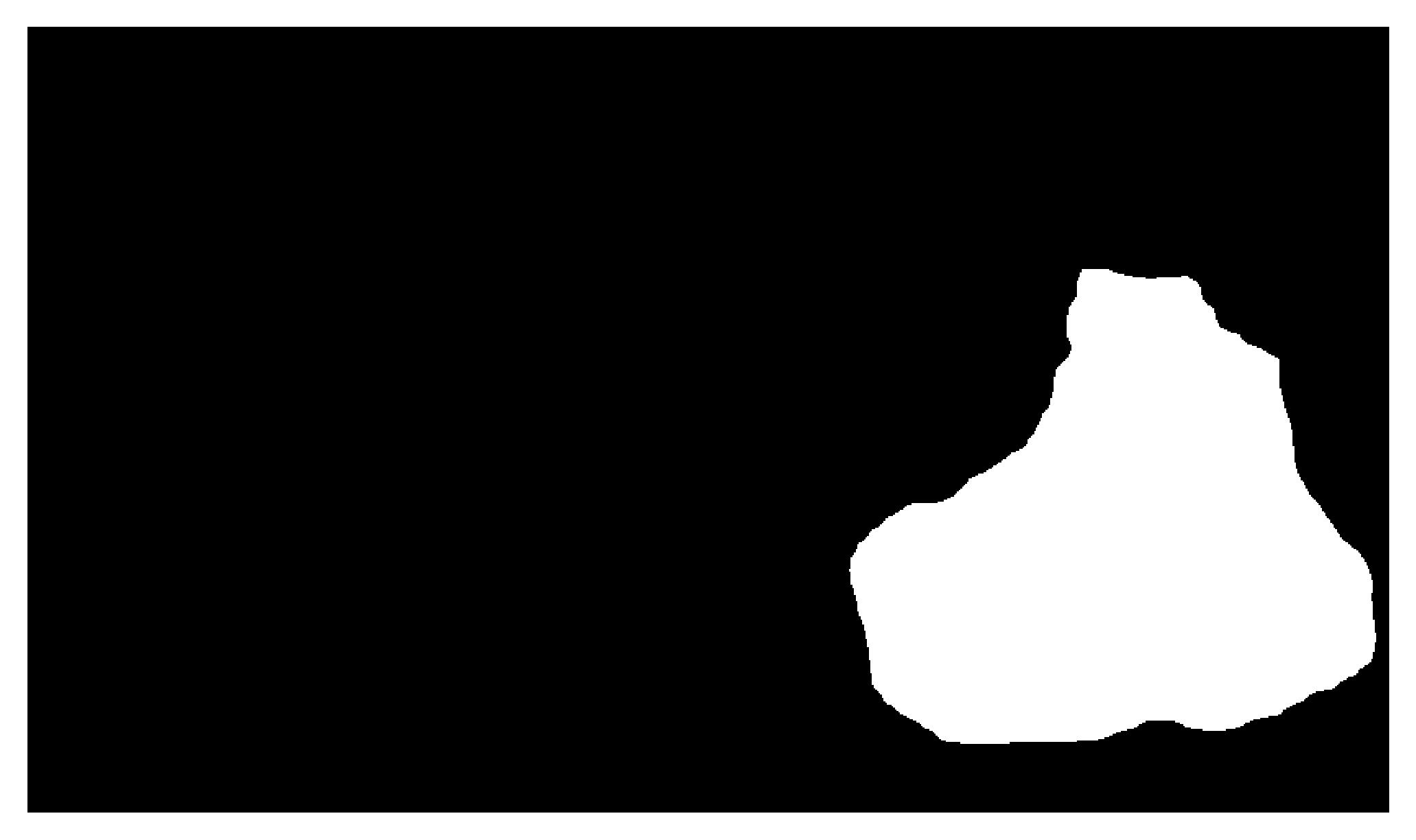}}
        \fbox{\includegraphics[width=0.1070\textwidth]{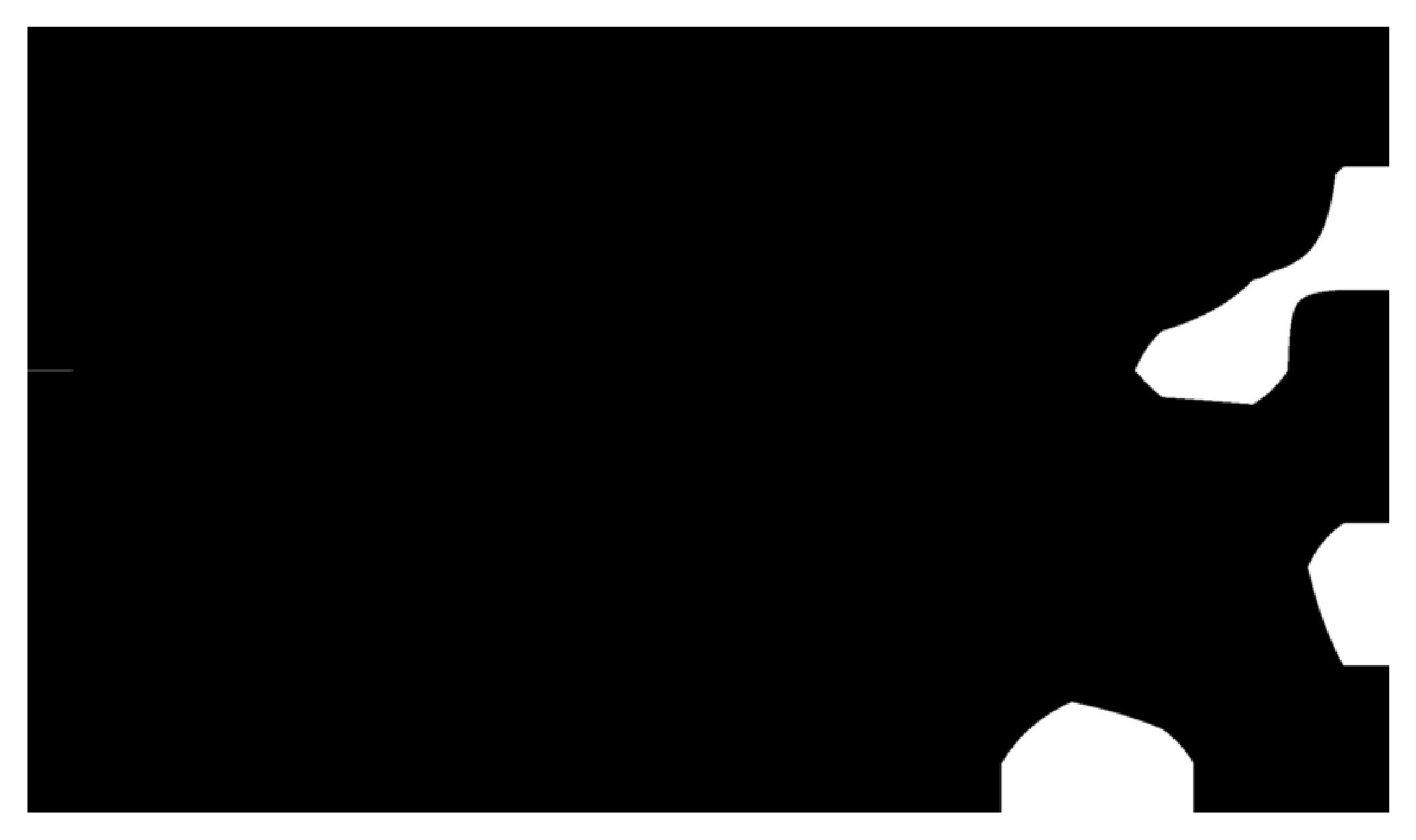}}
        \fbox{\includegraphics[width=0.1070\textwidth]{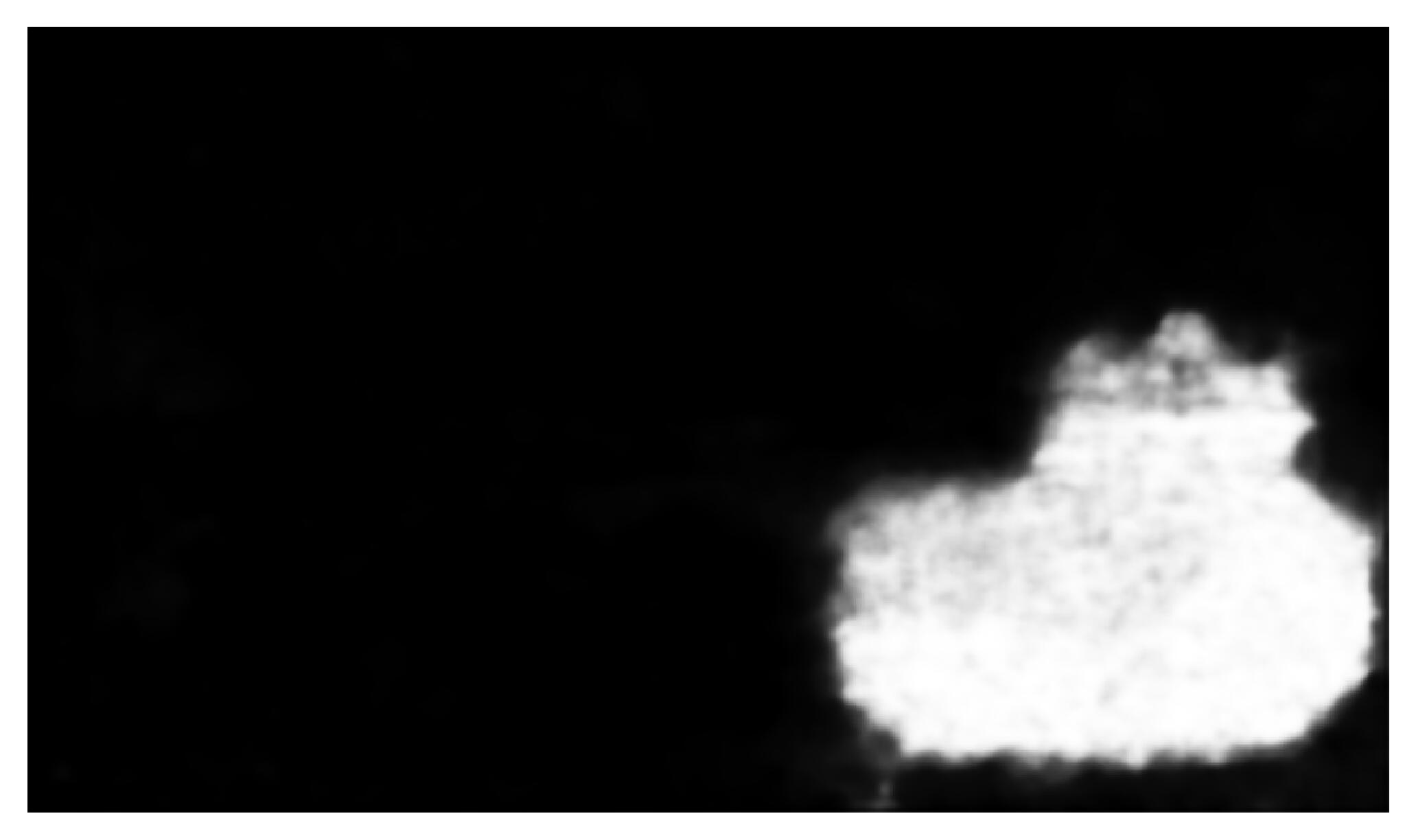}}
        \fbox{\includegraphics[width=0.1070\textwidth]{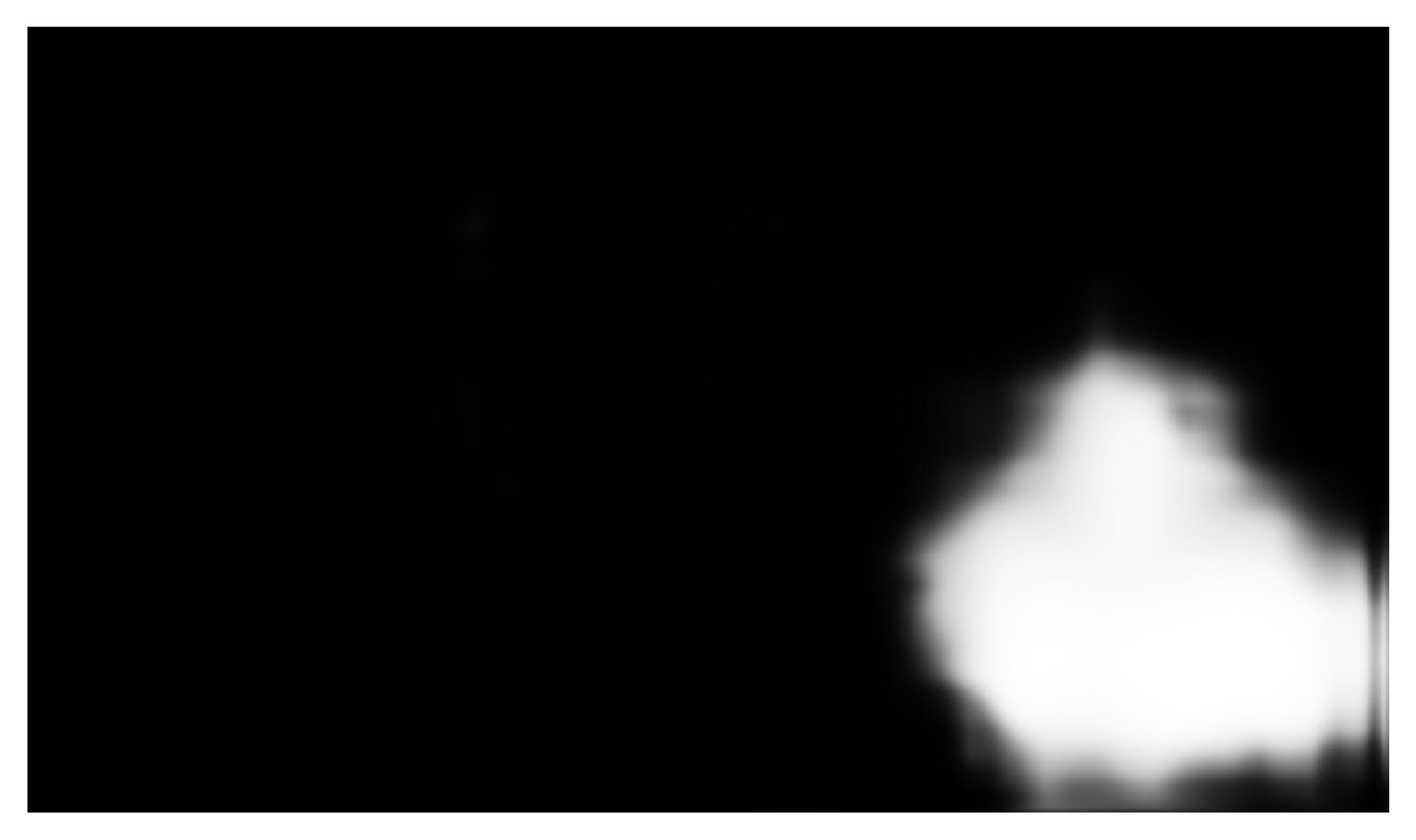}}
        \fbox{\includegraphics[width=0.1070\textwidth]{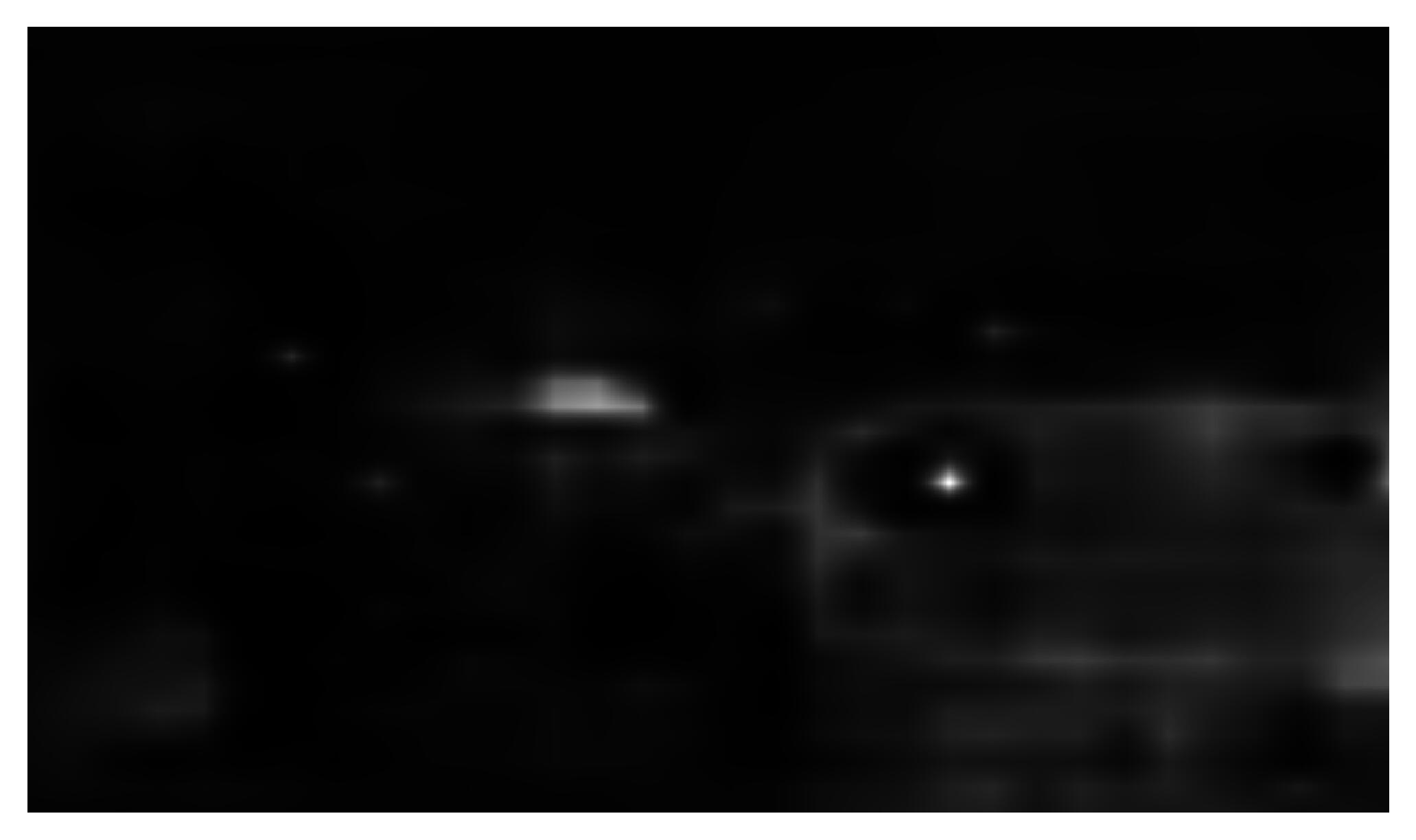}}
        \fbox{\includegraphics[width=0.1070\textwidth]{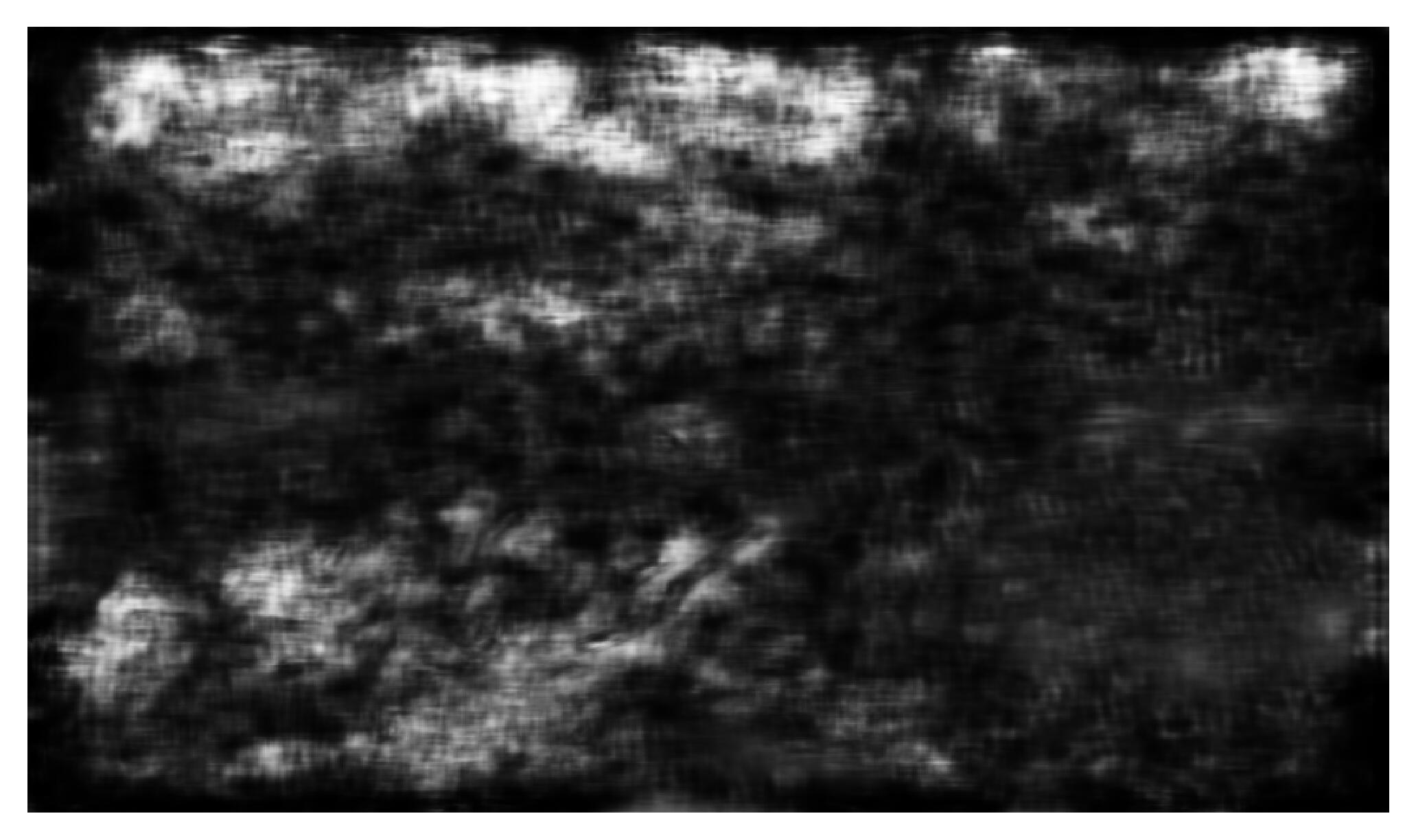}}
        \fbox{\includegraphics[width=0.1070\textwidth]{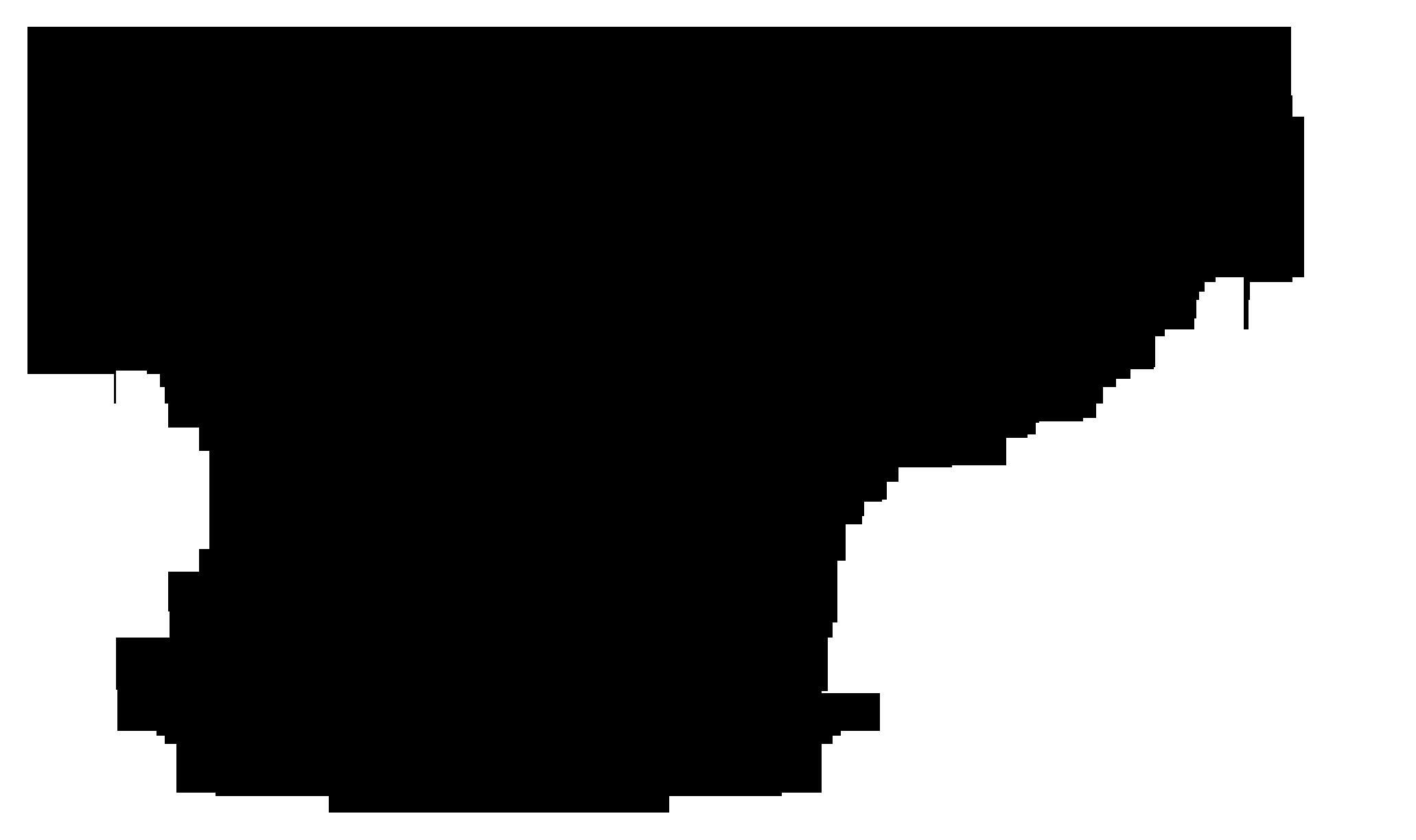}}
    \end{minipage}

	\vspace*{-0.1\baselineskip}

    \begin{minipage}[t]{1\textwidth}
    	\setlength{\fboxrule}{-2pt}
        \makebox[0.070\textwidth][r]{\raisebox{8pt}{\smallerr Splicing\hspace{2pt}}}
        \fbox{\includegraphics[width=0.1070\textwidth]{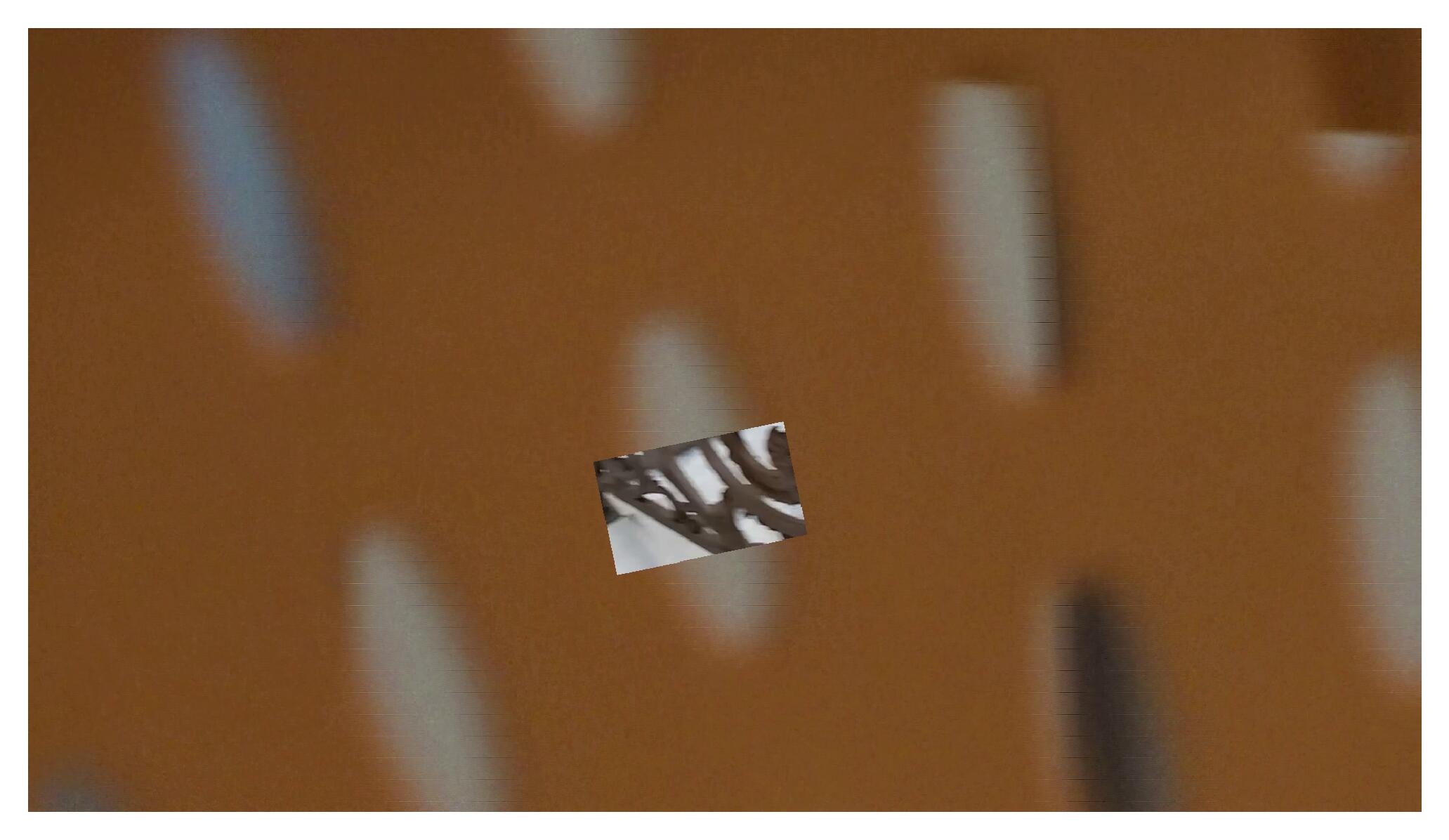}}
        \fbox{\includegraphics[width=0.1070\textwidth]{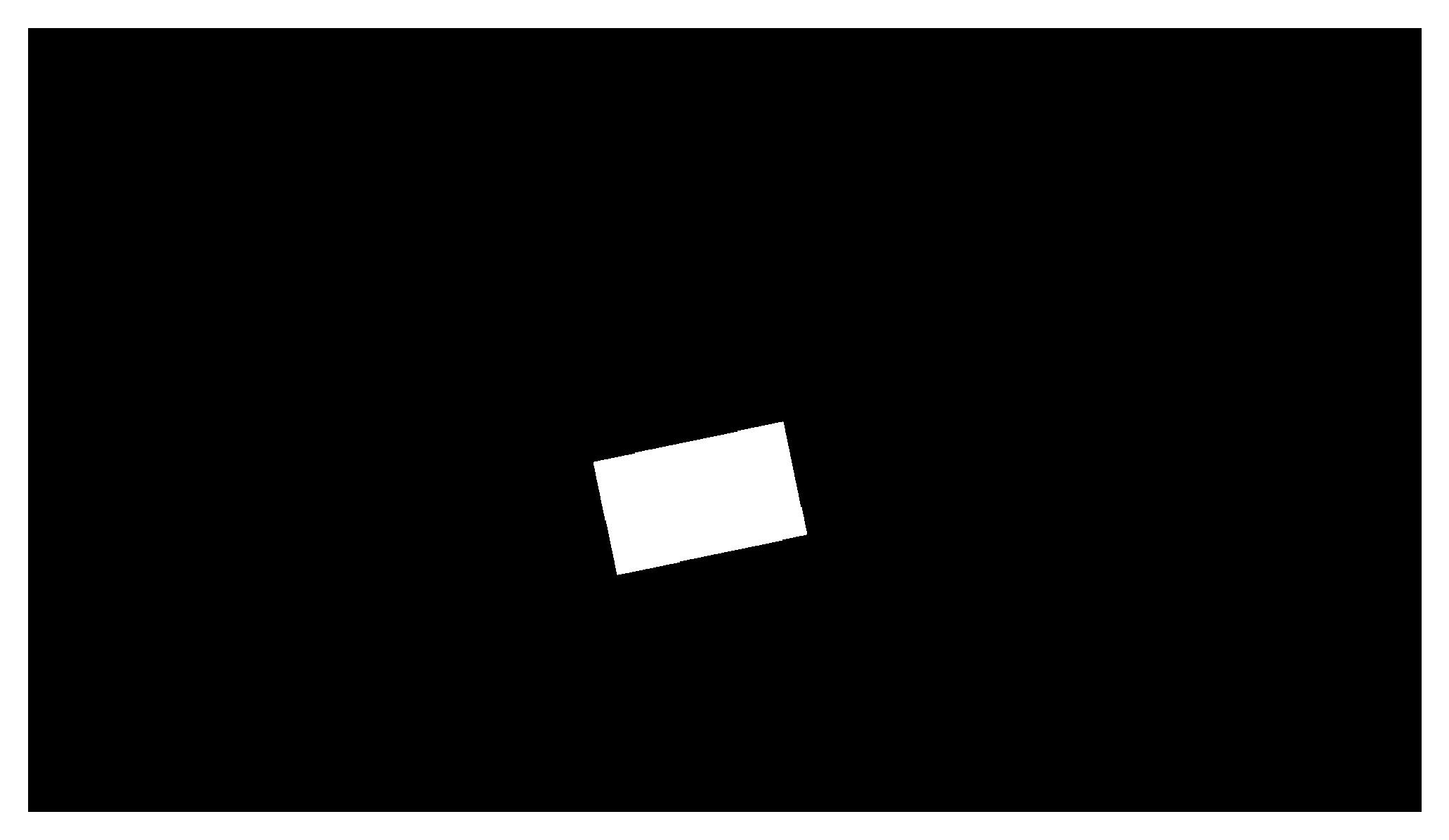}}
        \fbox{\includegraphics[width=0.1070\textwidth]{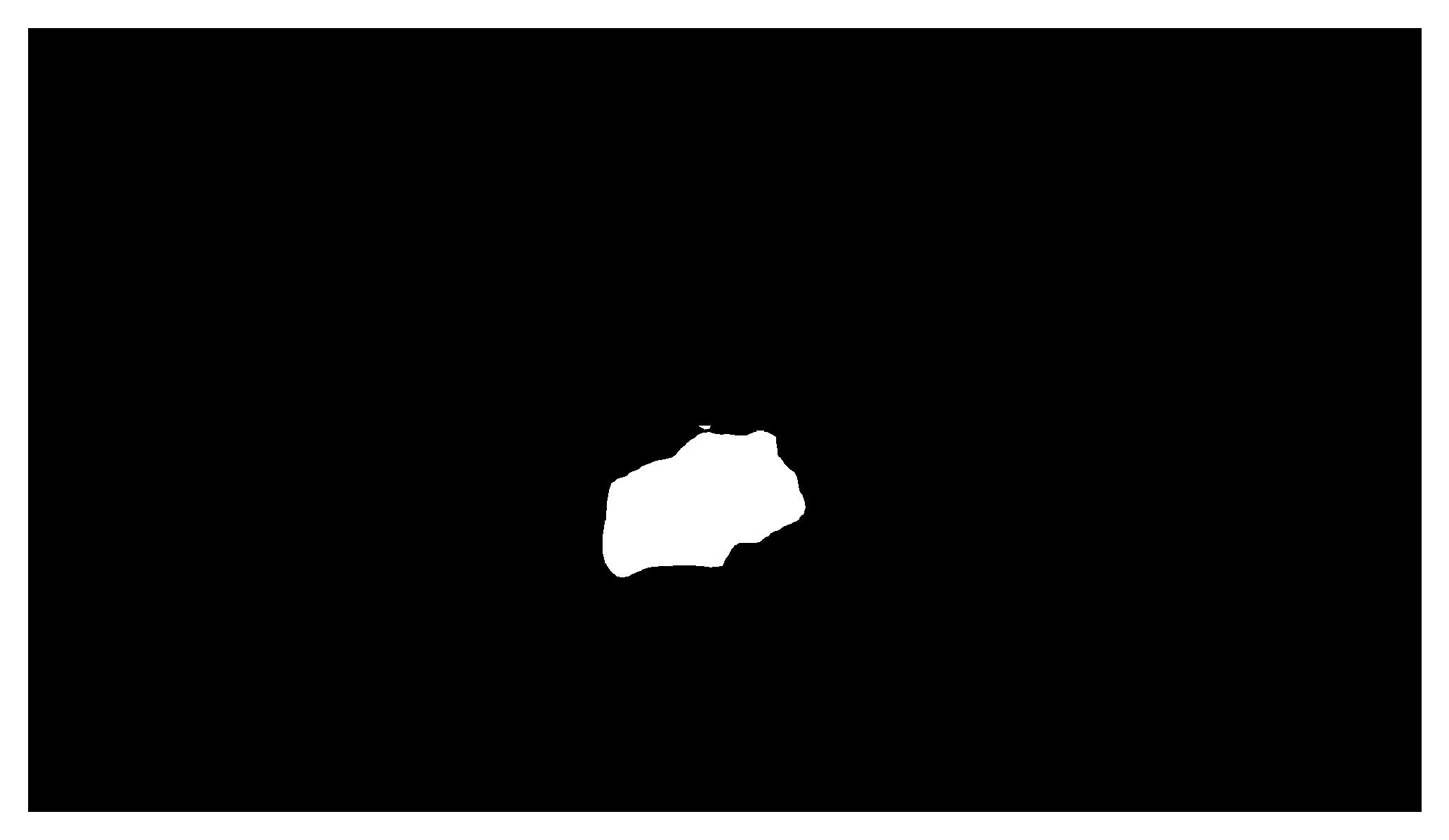}}
        \fbox{\includegraphics[width=0.1070\textwidth]{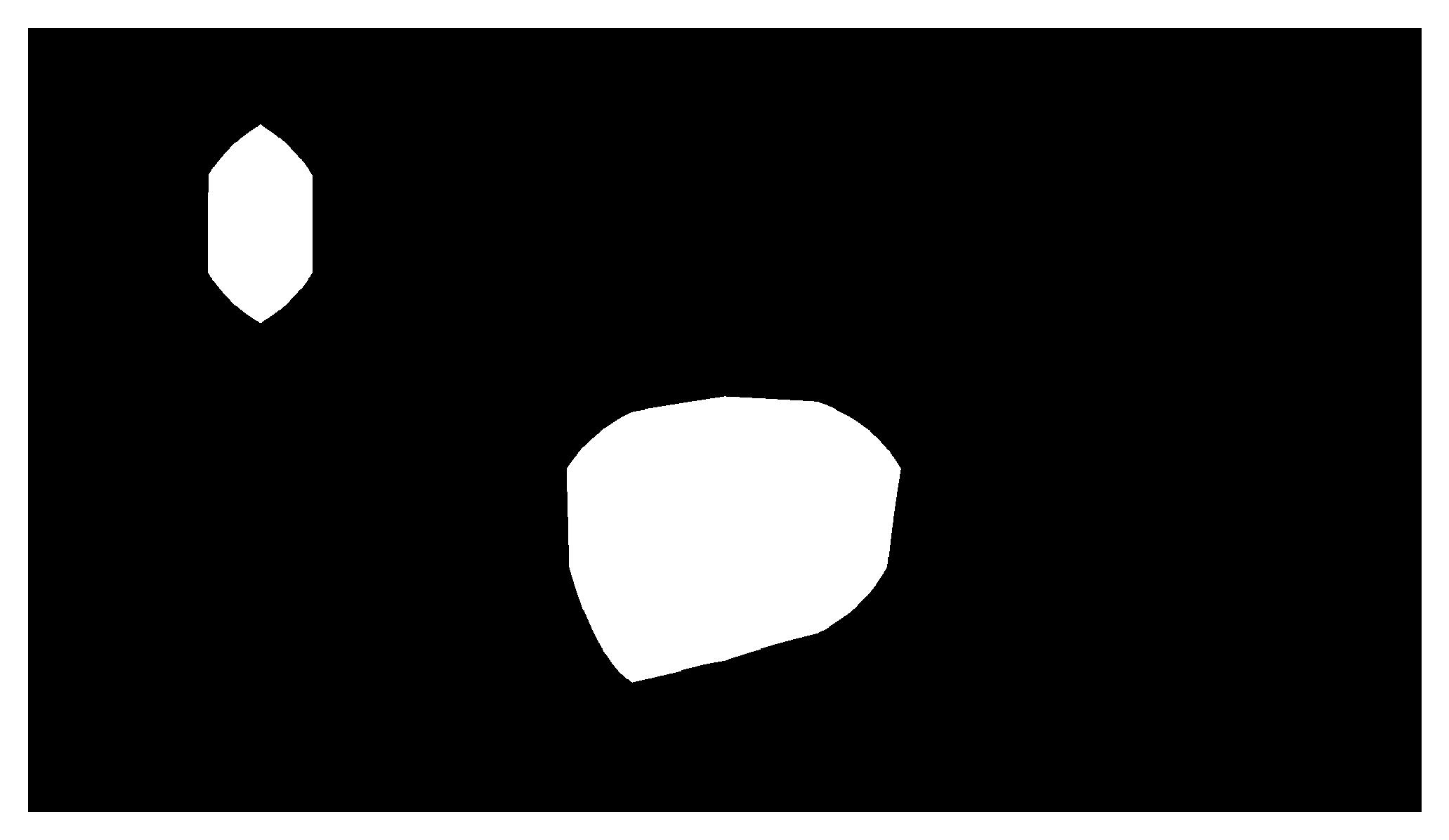}}
        \fbox{\includegraphics[width=0.1070\textwidth]{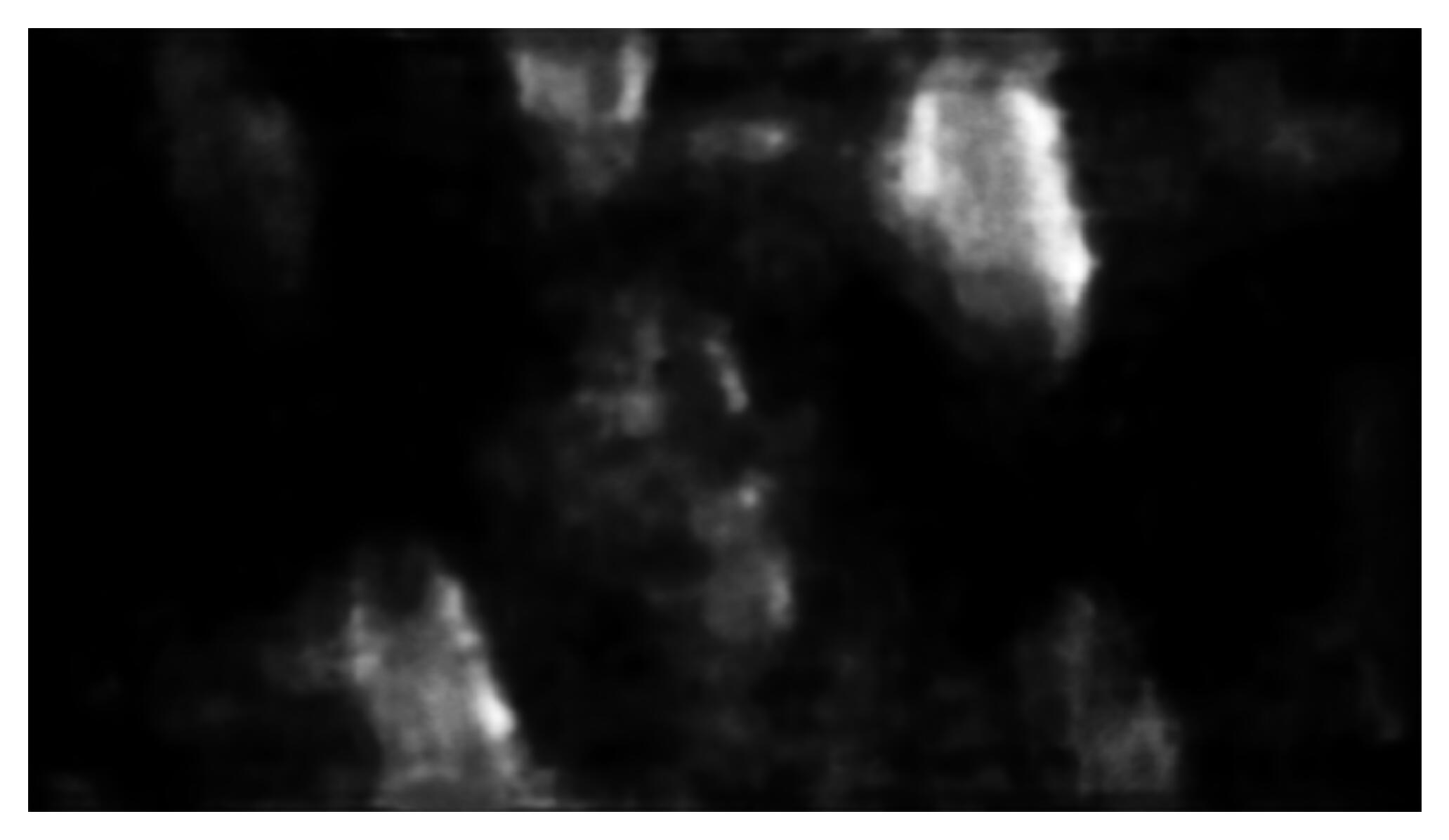}}
        \fbox{\includegraphics[width=0.1070\textwidth]{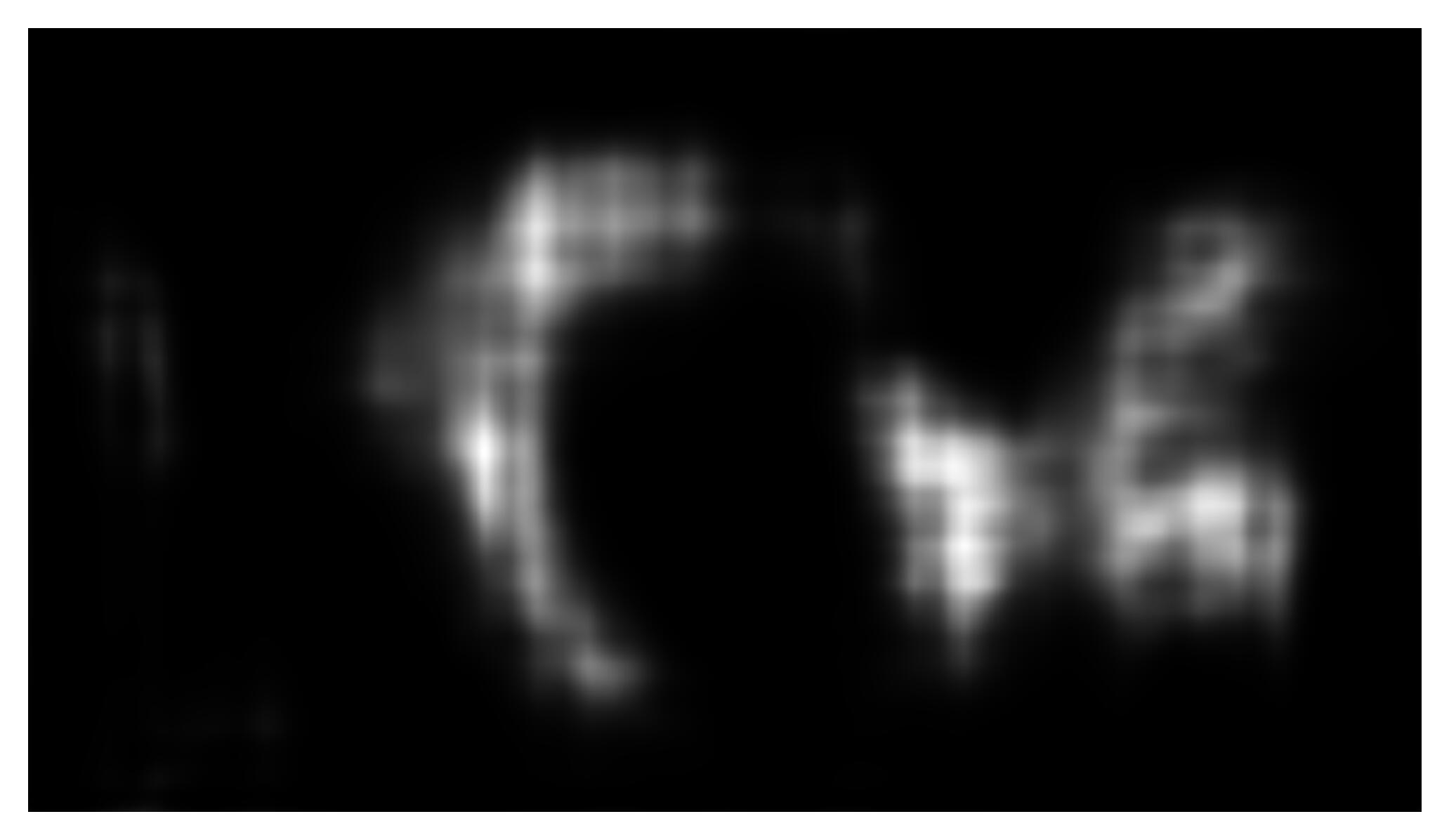}}
        \fbox{\includegraphics[width=0.1070\textwidth]{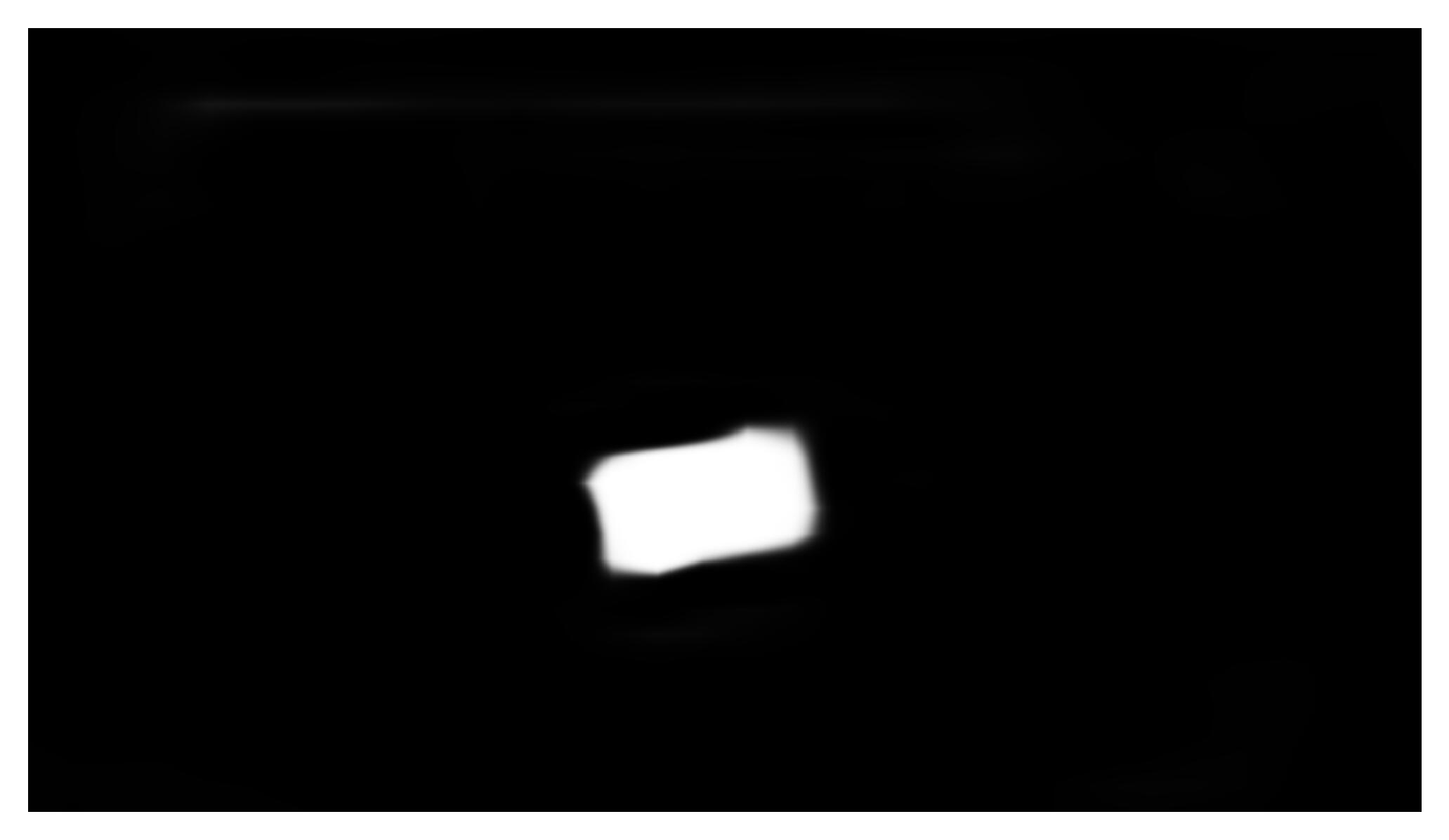}}
        \fbox{\includegraphics[width=0.1070\textwidth]{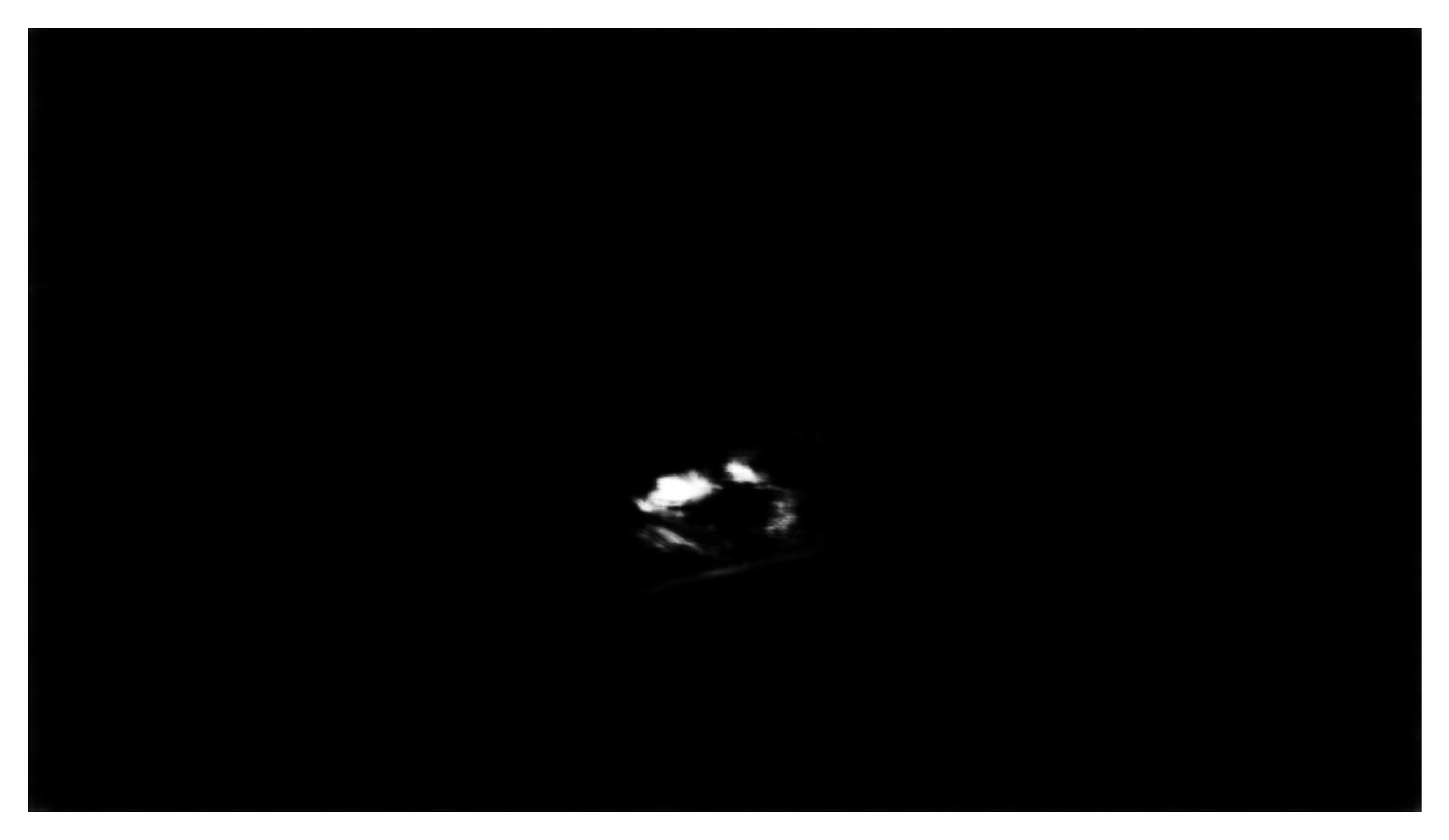}}
        \fbox{\includegraphics[width=0.1070\textwidth]{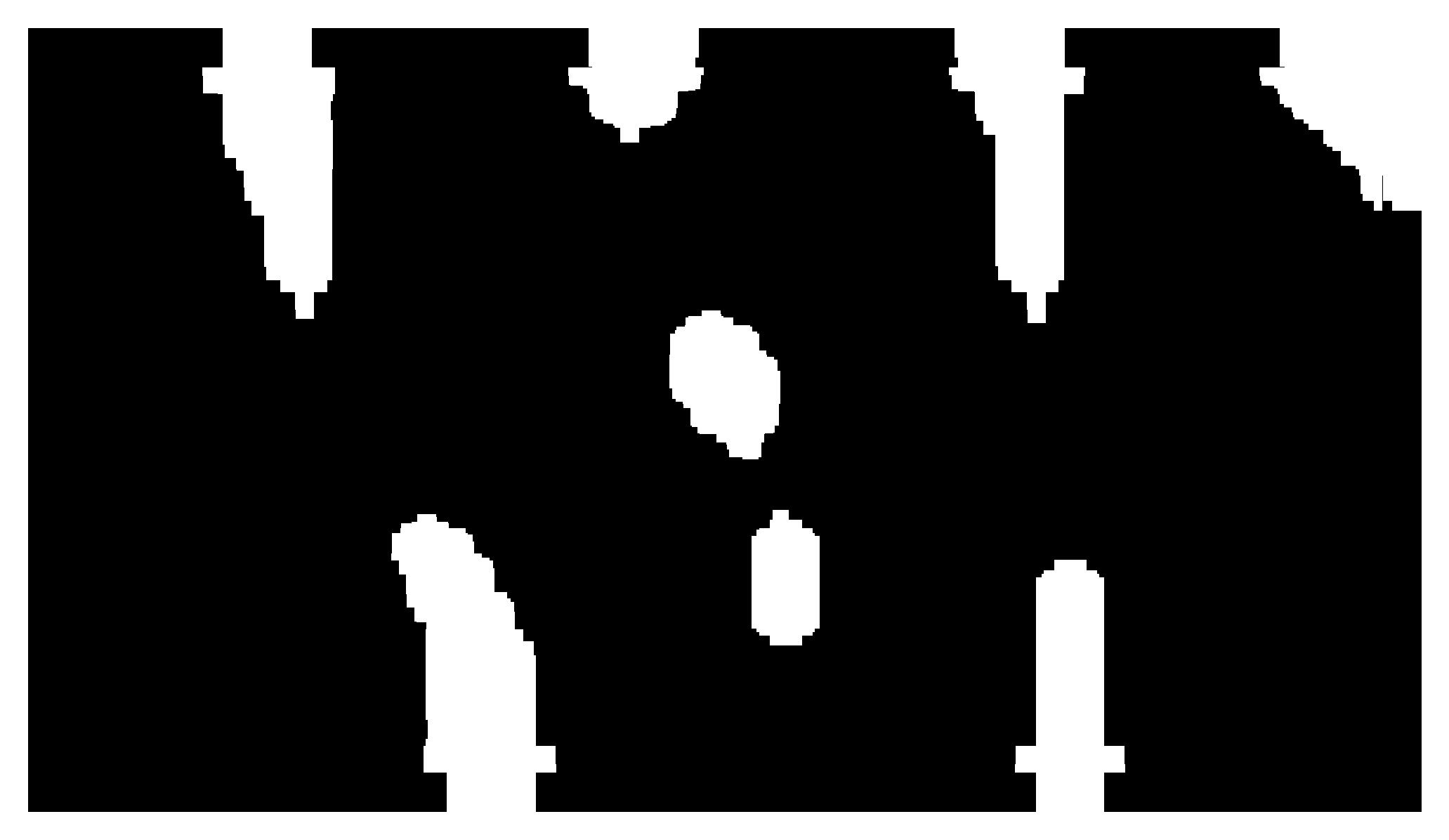}}
    \end{minipage}

	\vspace*{-0.1\baselineskip}

    \begin{minipage}[t]{1\textwidth}
    	\setlength{\fboxrule}{-2pt}
        \makebox[0.070\textwidth][r]{\raisebox{8pt}{\smallerr Editing\hspace{2pt}}}
        \fbox{\includegraphics[width=0.1070\textwidth]{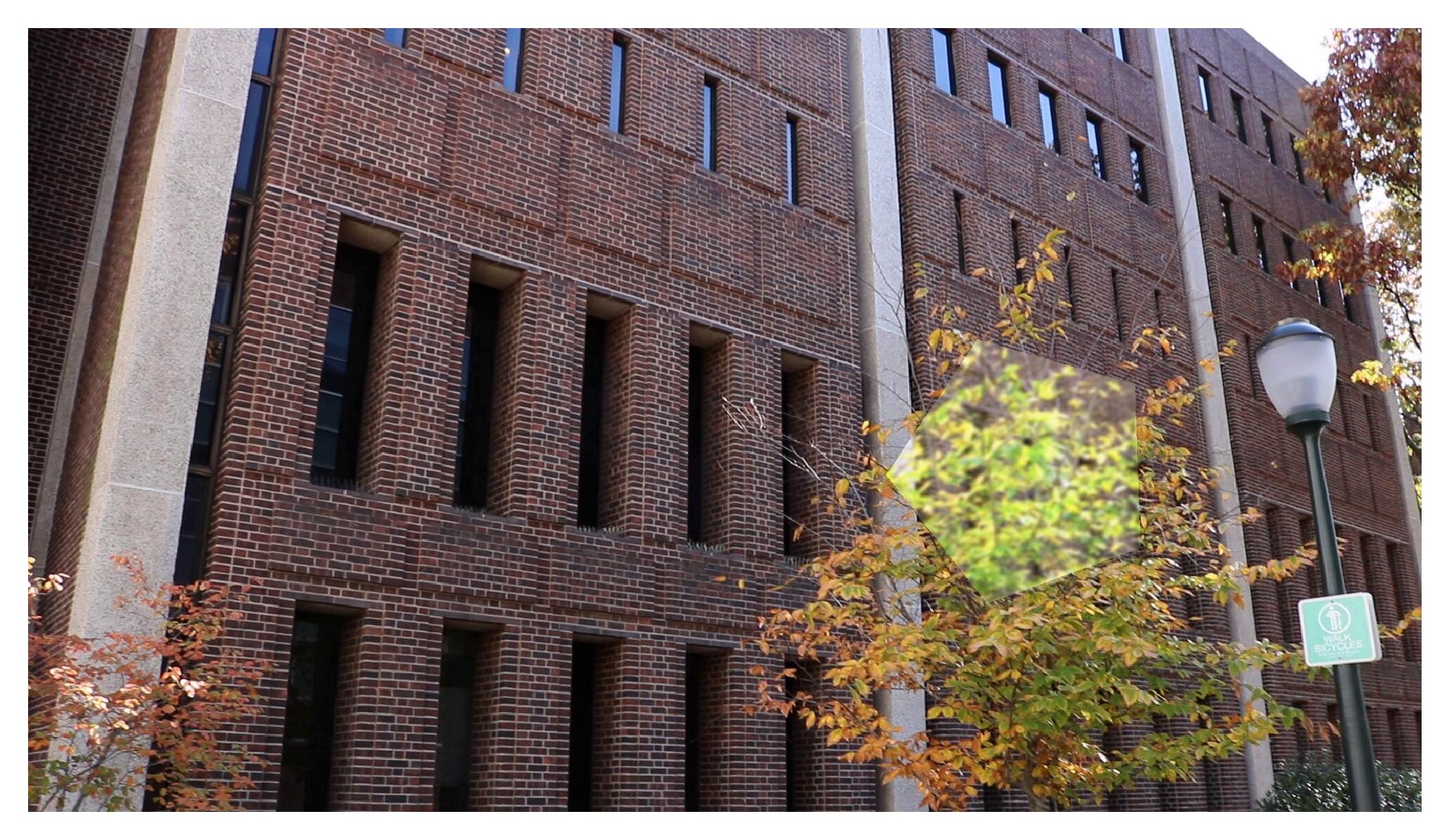}}
        \fbox{\includegraphics[width=0.1070\textwidth]{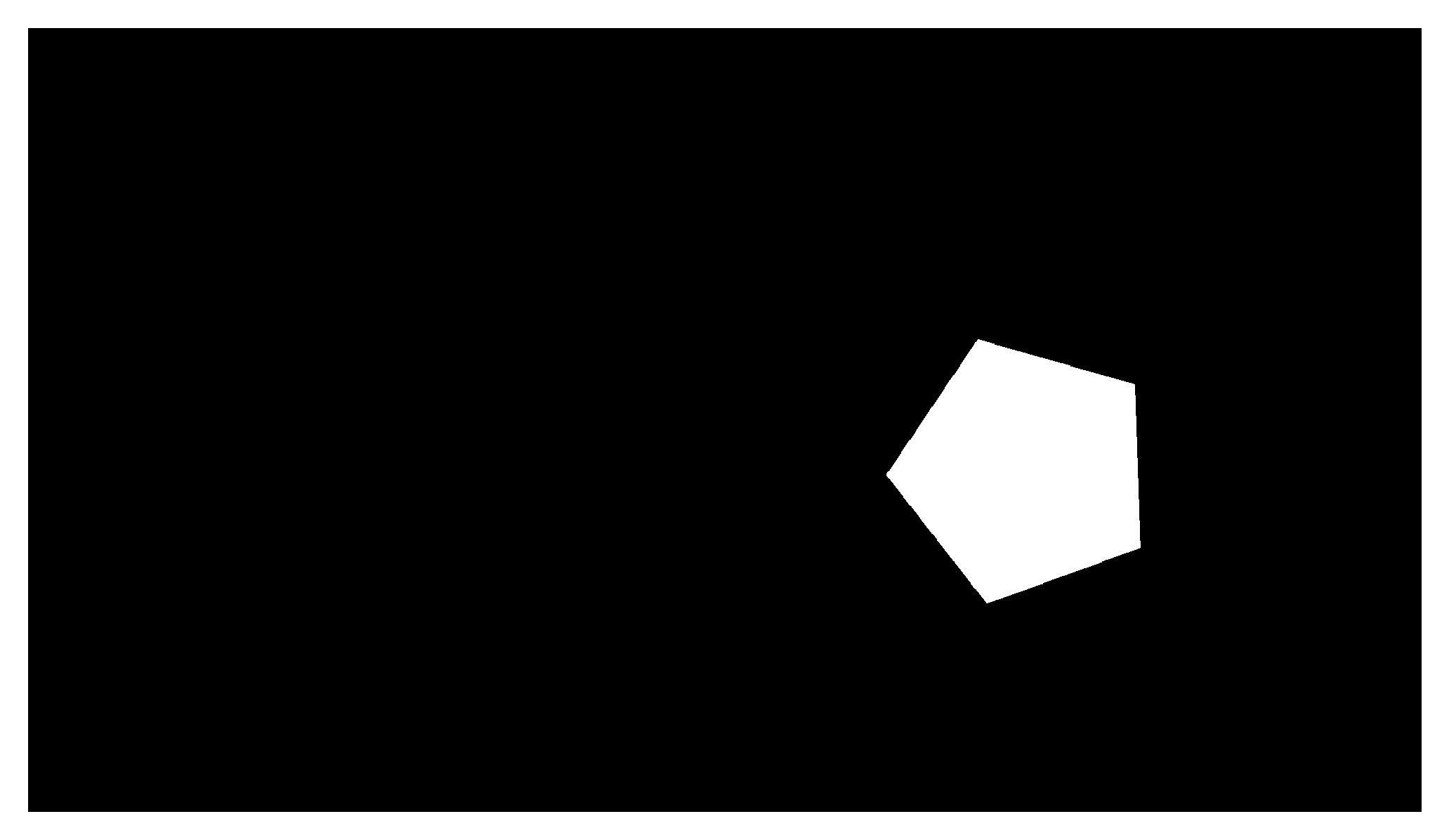}}
        \fbox{\includegraphics[width=0.1070\textwidth]{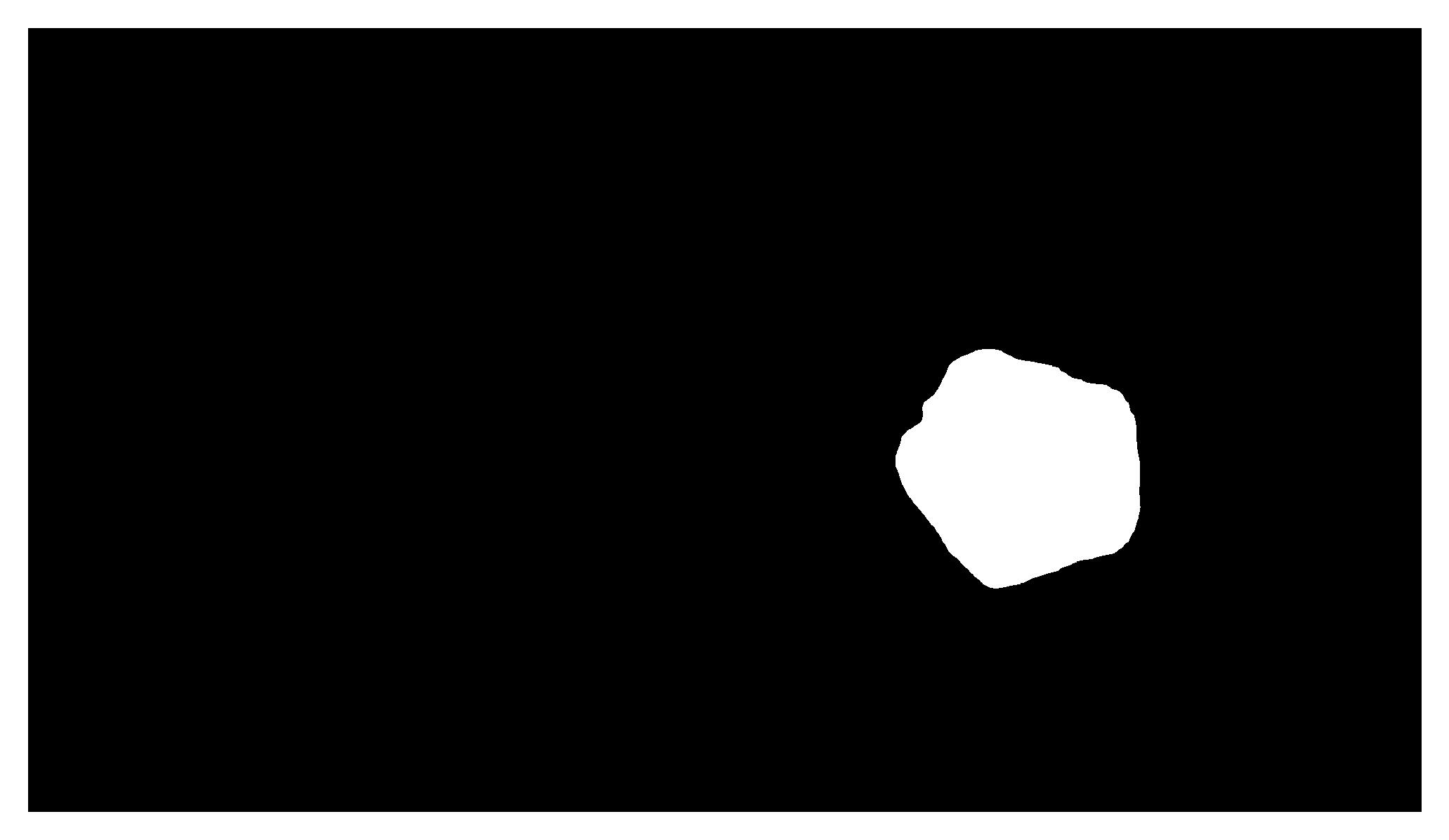}}
        \fbox{\includegraphics[width=0.1070\textwidth]{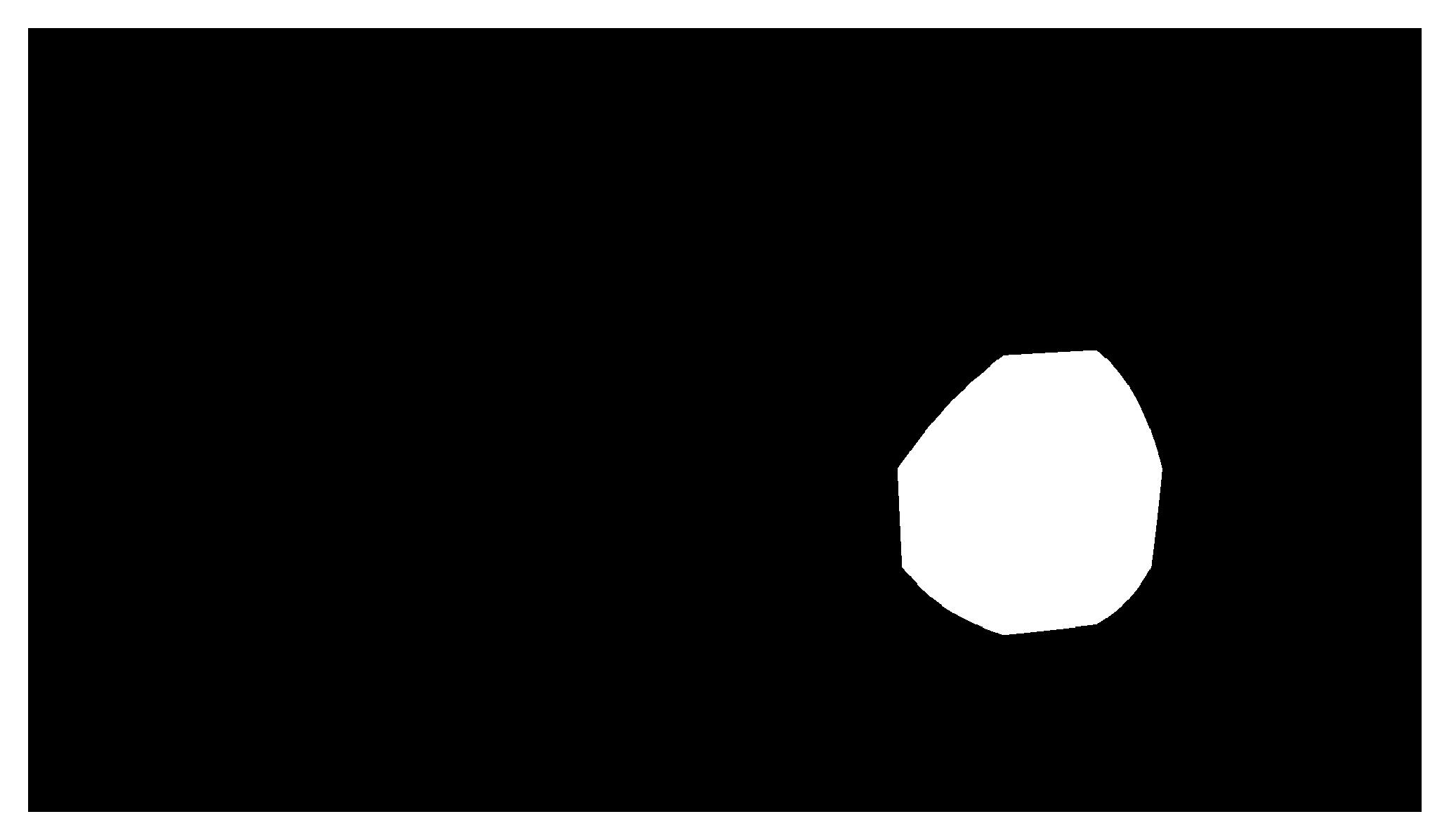}}
        \fbox{\includegraphics[width=0.1070\textwidth]{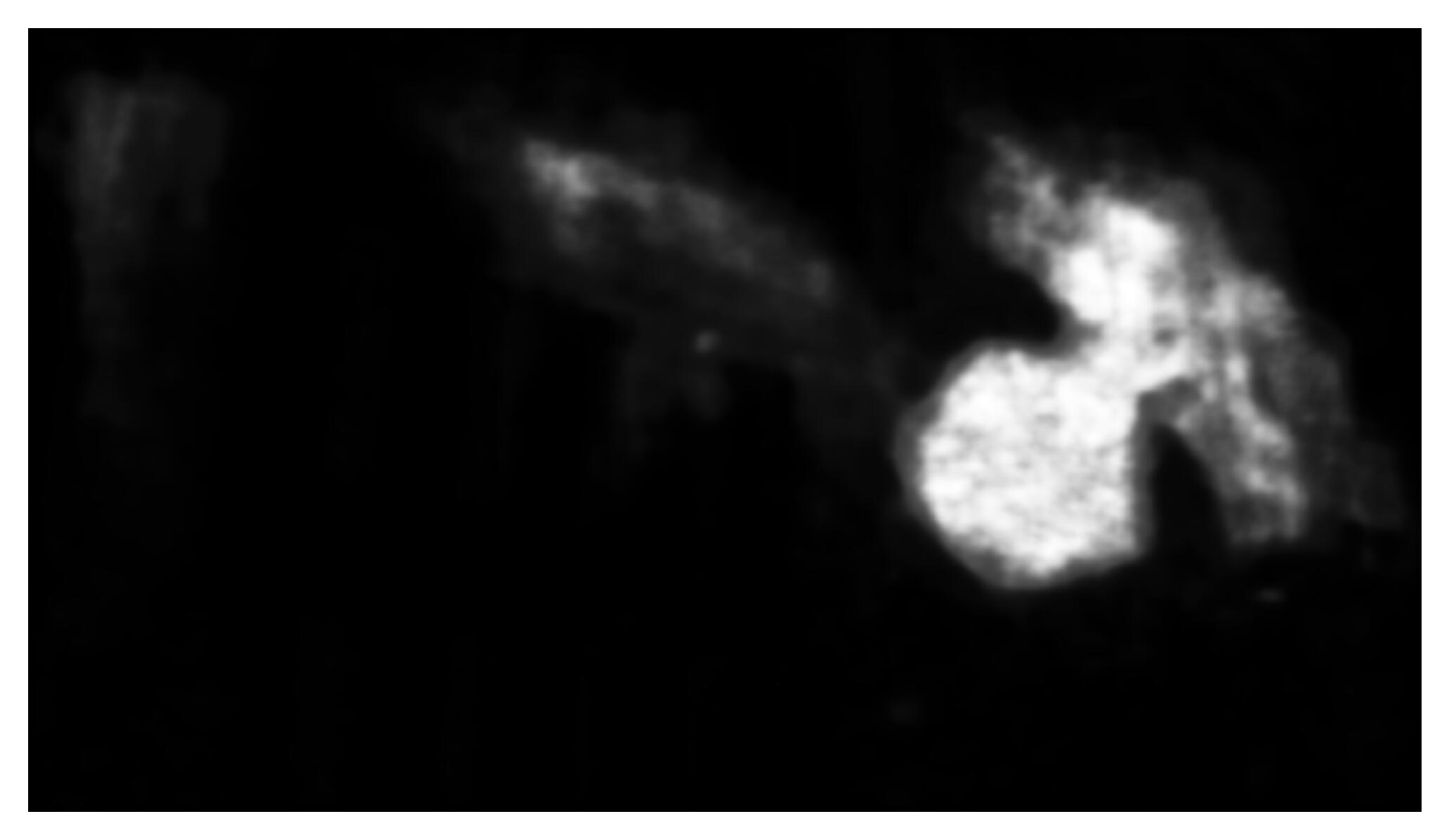}}
        \fbox{\includegraphics[width=0.1070\textwidth]{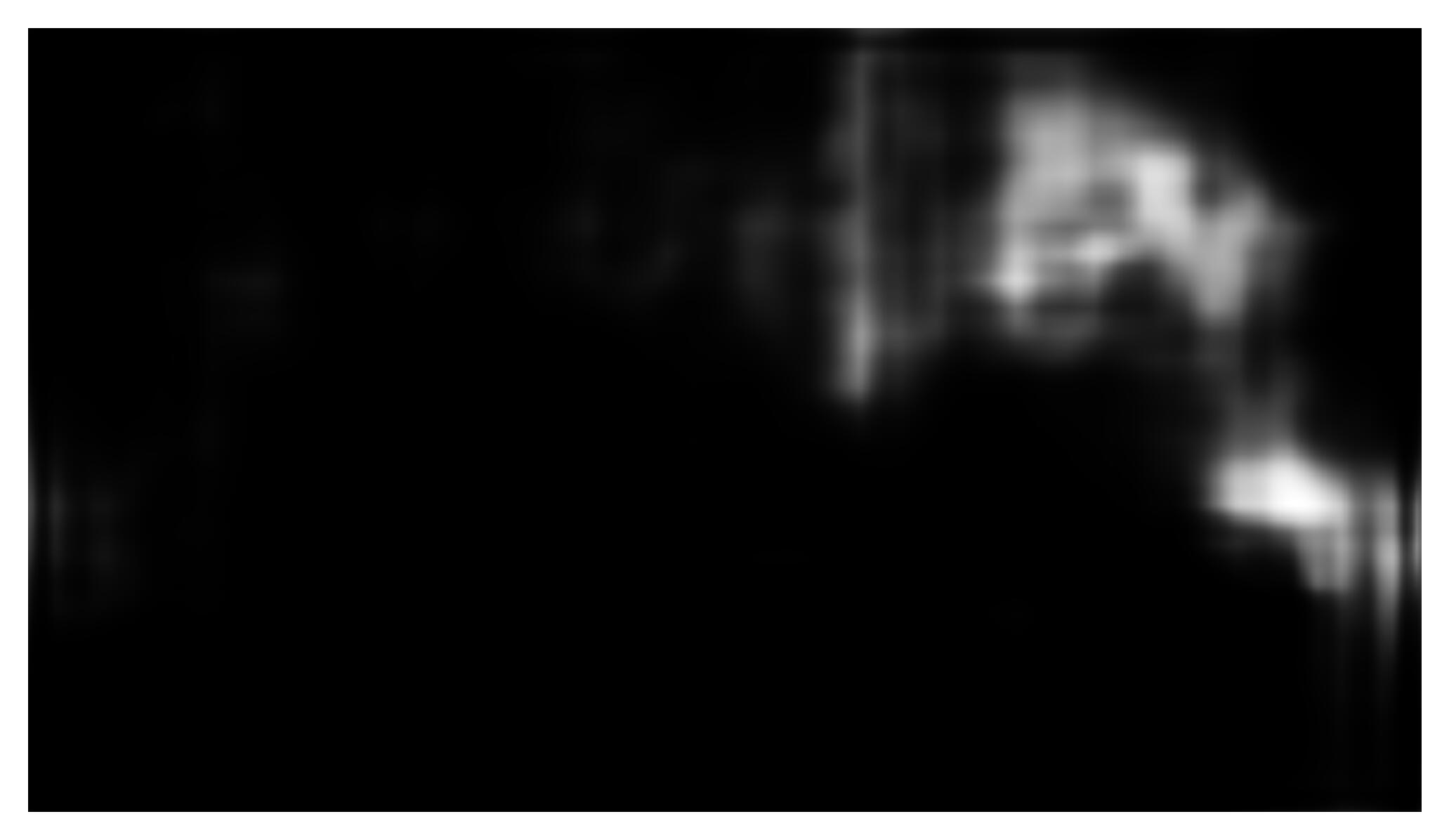}}
        \fbox{\includegraphics[width=0.1070\textwidth]{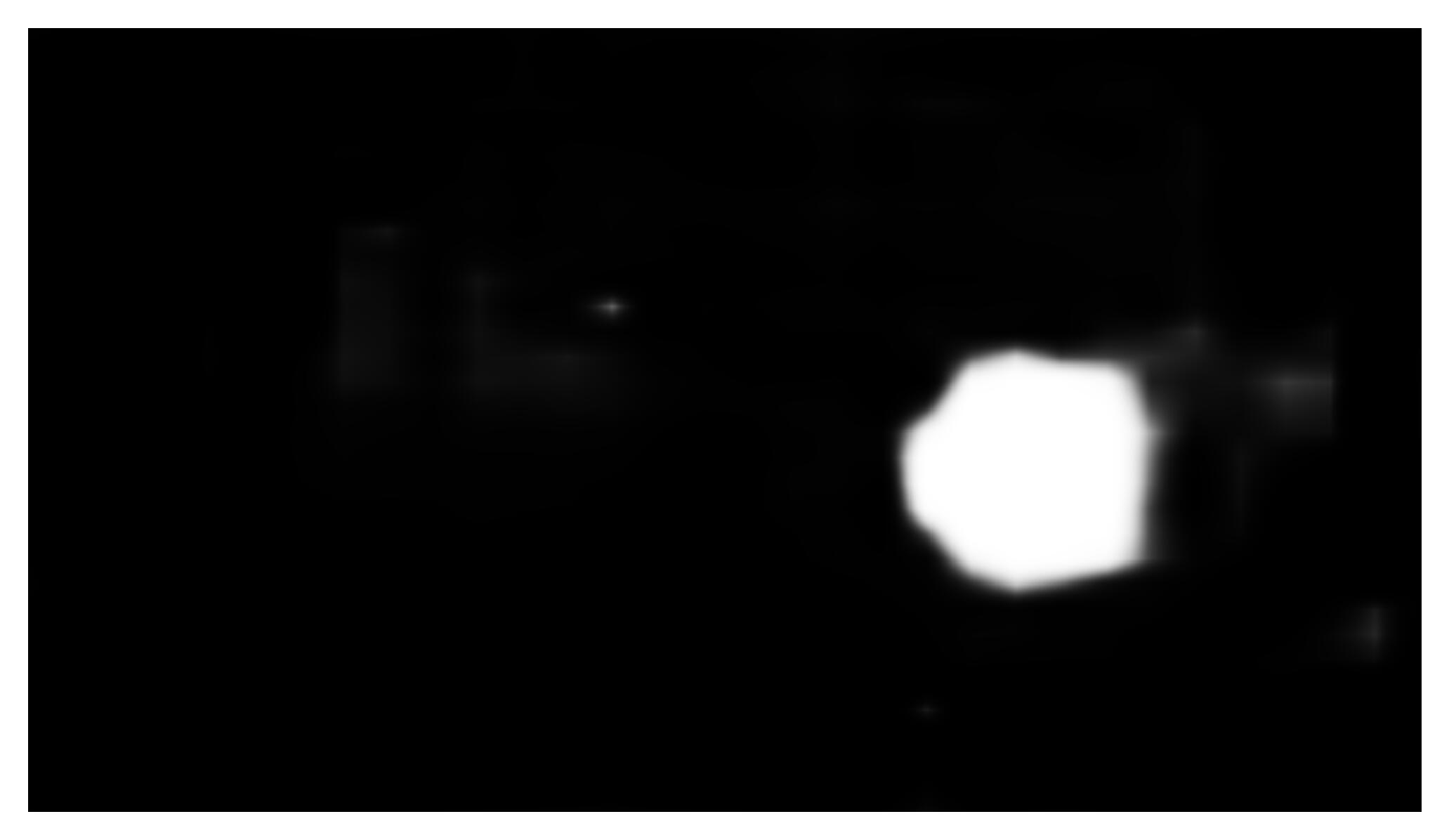}}
        \fbox{\includegraphics[width=0.1070\textwidth]{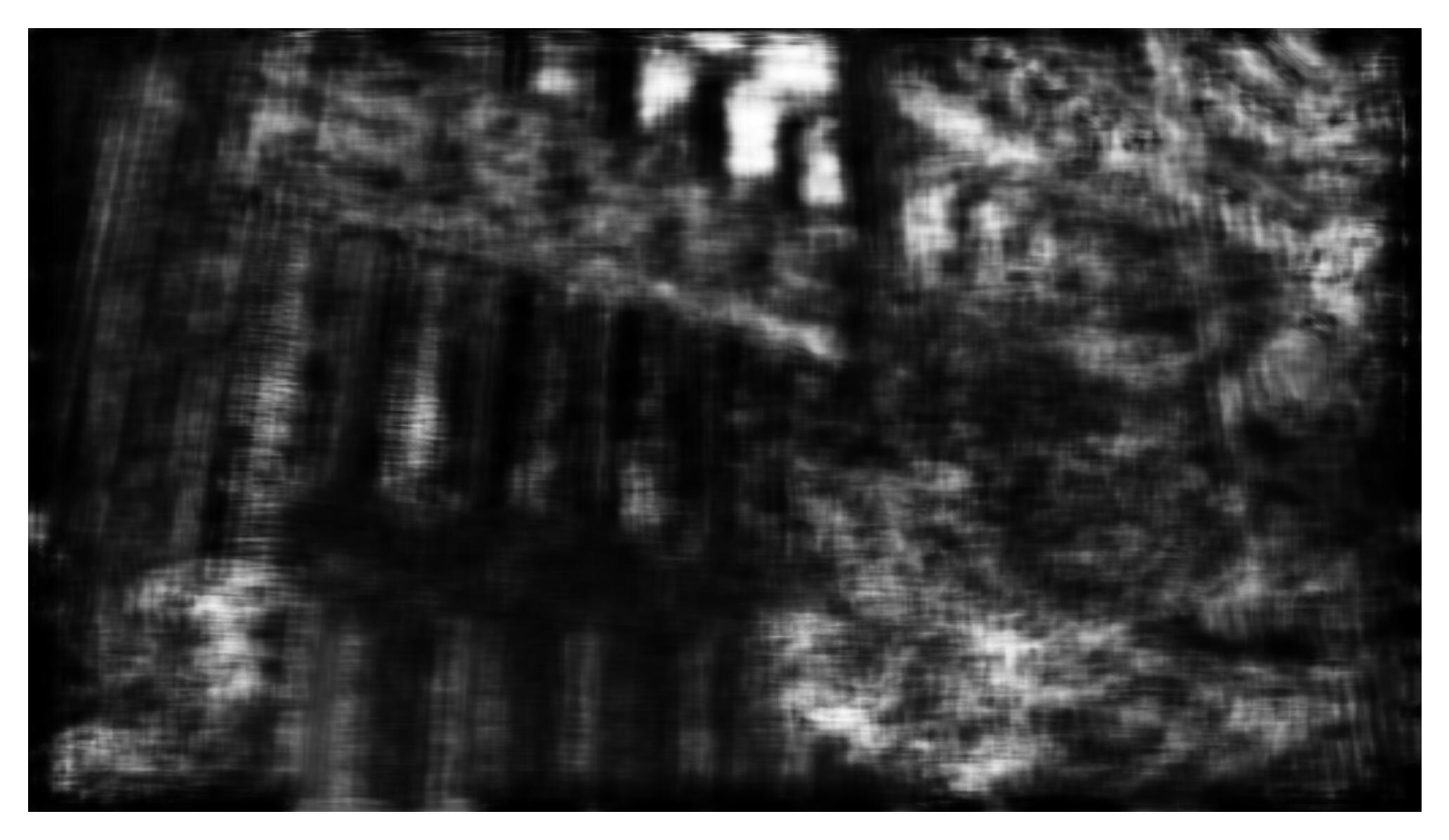}}
        \fbox{\includegraphics[width=0.1070\textwidth]{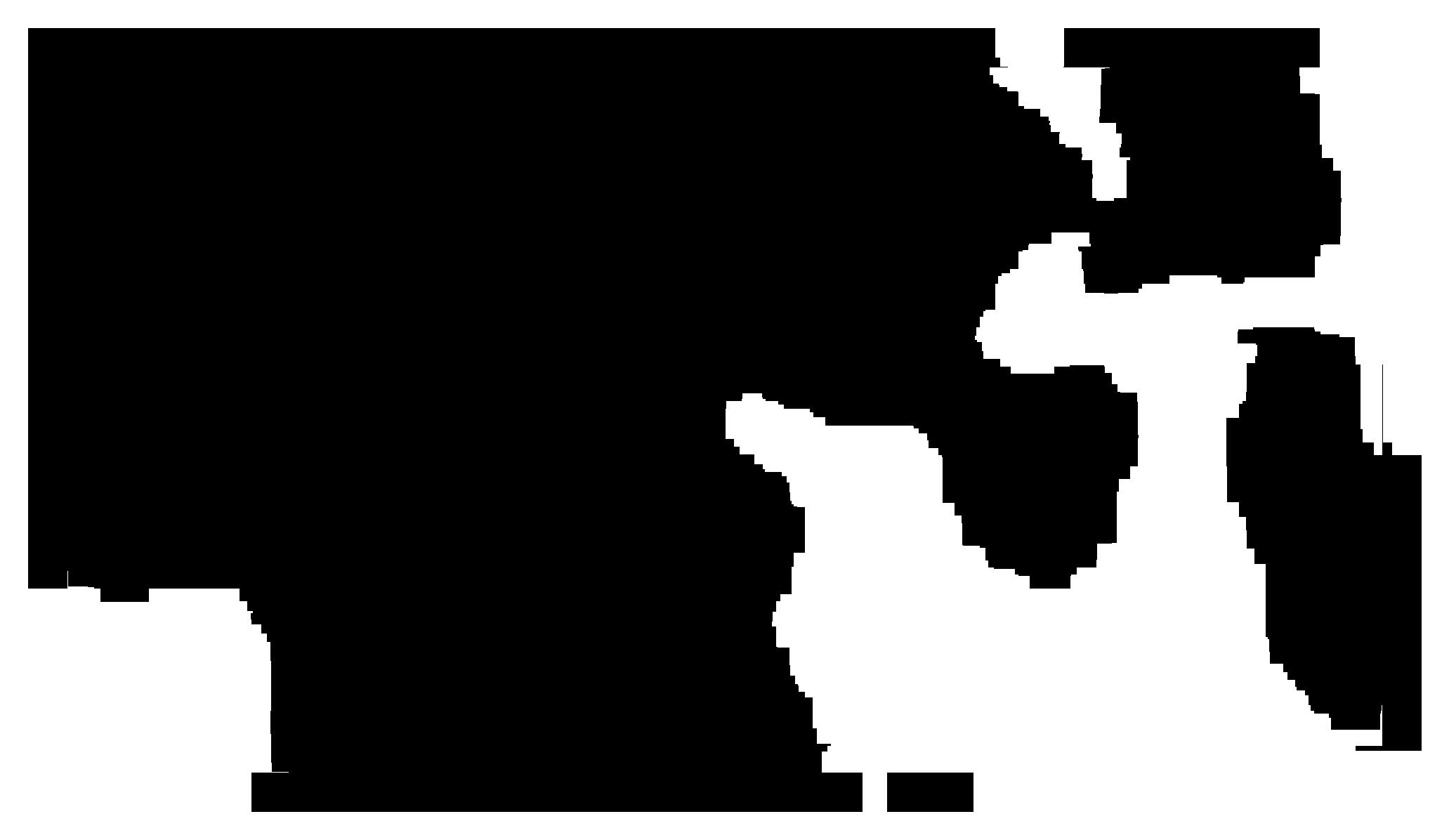}}
    \end{minipage}

    \caption{Localization results from our proposed network as well as VideoFACT~\cite{VideoFACT}, VIDNet~\cite{VIDNet}, DVIL~\cite{DVIL}, MVSS-Net~\cite{MVSS-Net}, ManTra-Net~\cite{ManTra-Net}, and FSG~\cite{FSG} on 4 different manipulation types in the \pooleddsabrv-IND dataset. We note that we do not provide localization results for deepfake detectors because these algorithms only perform detection.}
	\label{fig:localization_results}

	\pulluppp\pullupp\pullup
\end{figure*}

%% file: sections/Data_v0.tex

\section{\pooleddsname Dataset}
\label{sec:uvfa}

\subsection{In-distribution Data (\pooleddsabrv-IND)}
\label{subsec:uvfa_ind}

To train and evaluate our network, we combined videos from 5 different publicly available video manipulation datasets across 4 types of forgery to create the in-distribution subset of \pooleddsabrv, denoted as \pooleddsabrv-IND. 

\subheader{Splicing and Editing}
The VideoFACT VCMS and VPVM datasets~\cite{VideoFACT} were used to include splicing and editing video forgeries. Even though the manipulations in these datasets are simple in nature, they served as an important source of training data and benchmark for our network.

\subheader{Deepfakes}
Deepfaked videos were obtained from the DeepFakeDetection~\cite{DFD} dataset as well as the Deepfakes (DF) and Face2Face (F2F) subsets of the FaceForensics++~\cite{FF++} dataset.
We note that other subsets of the FaceForensics++ are created using different generators. We later use these subsets for out-of-distribution data evaluation.

\subheader{Inpainting} The Diagnostic Evaluation Benchmark for Video Inpainting (DEVIL) dataset~\cite{DEVIL} was used to include inpainted videos, which were created using state-of-the-art inpainting algorithms~\cite{JointOpt, DFCNet, CPNet, OPN, STTN, FGVC}.

\subsection{Out-of-distribution Data (\pooleddsabrv-OOD)}
\label{subsec:uvfa_ood}
To thoroughly assess forensic generalizability, we gathered data from 8 new sources to
created a separate out-of-distribution subset, \pooleddsabrv-OOD (shown in Tab.~\ref{table:dataset_statistics}).
This subset contains new manipulations and distinct content not seen in training. We note that the same broad families of manipulations remained consistent (i.e. inpainting, deepfakes, editing, splicing), but different, and often more advanced, algorithms were used to implement these.

\subheader{Splicing and Editing}
We created the DAVIS Standard Manipulation dataset to test our approach's performance on more sophisticated splicing and editing forgeries. Moreover, these manipulations are performed on distinct content distribution and resolution than our training data. To make this dataset, we spliced or edited objects specified by the segmentation masks in the DAVIS dataset~\cite{DAVIS}..

\subheader {Deepfakes}
We included videos from the FaceSwap and NeuralTextures subsets of FF++~\cite{FF++}, as well as the FaceShifter dataset~\cite{FaceShifter}. Each of these datasets includes deepfaking techniques not seen during training.

\subheader{Inpainting}
We included the DAVIS Inpainting dataset~\cite{VideoFACT} to evaluate our method on detecting new inpainting algorithms. This dataset was made by applying state-of-the-art techniques E2FGVI-HQ~\cite{E2FGVI} \& FuseFormer~\cite{FuseFormer} on objects specified by the masks provided in the DAVIS dataset~\cite{DAVIS}.

\subsection{VideoSham Dataset}
\label{subsec:videosham}
To test our approach's ability to detect and localize new manipulation types not seen during training, we included the VideoSham~\cite{VideoSham} dataset in our experiments.
VideoSham is a highly challenging dataset that
contains professionally edited videos, made using a diverse combination of unknown advanced techniques for better credibility.

%% file: sections/Experiments_v1.tex
\pullupp
\section{Experiments}
\label{sec:experiments}

\subsection{Experimental Setup}
\label{subsec:experimental_setup}

\subheader{Training}
We pretrained the spatial forensic residual module on the Video-ACID dataset~\cite{Video-ACID}, followed by full network training on the datasets listed in Tab. 1. Loss parameters $\gamma$, $\alpha$, and $\beta$ were adjusted dynamically per epoch (details in supplementary material).

\subheader{Testing}
We comprehensively tested UVFNet in two settings: (1) multi-manipulation, where videos contained one of several unknown manipulation types (e.g., deepfake, inpainting, splicing, editing), and (2) single-manipulation, focusing on a single known forgery type.

\subheader{Metrics}
To evaluate all methods, we used frame-level detection mAP and pixel-level localization F1 scores. For methods that do not offer localization, only detection scores are reported. For non-detection methods, frame-level detection scores are derived from the highest per-pixel detection probability, a common approach in prior works~\cite{DVIL, VIDNet, MVSS-Net, TruFor, ManTra-Net, FSG} to ensure a fair comparison.

\subheader{Competing Methods}
We benchmarked \ournetworkname against 9 state-of-the-art systems, categorized as:
(1) Splicing/Editing: MVSS-Net~\cite{MVSS-Net}, MantraNet~\cite{ManTra-Net}, FSG~\cite{FSG};
(2) Inpainting: VIDNet~\cite{VIDNet}, DVIL~\cite{DVIL};
(3) Deepfake: SBI~\cite{SBI}, MADD~\cite{MADD}, CE.ViT~\cite{CE.ViT};
and (4) General: VideoFACT~\cite{VideoFACT}.
For a fair comparison, we retrained the top-performing methods of each group using the exact same training data as ours(see supplemental materials). We excluded deepfake detectors in this process because they need to isolate and operate on a face, which is impossible in the multi-manipulation setting. Nonetheless, these systems have all seen similar deepfake data as ours.
We also compared against an ensemble of the retrained best methods to show that \textit{a naive combination of forensic modalities is insufficient to reliably detect multiple forms of forgery}.


\input{tables/pooled_results}
\pullupp
\subsection{Results for Multi-Manipulation Experiment}
\label{subsec:mult_manip_eval}

In this experiment, we evaluated our ability to detect and localize fake content in videos that are falsified with one of many possible manipulation types (deepfake, inpainting, splicing, editing), and compared our performance against competing specialized and general detectors. Our testing data is split into 3 categories:
1) in-distribution data (\pooleddsabrv-IND) with content distribution and manipulation techniques similar to the training set,
2) out-of-distribution data (\pooleddsabrv-OOD) with unseen content distribution and different manipulation algorithms (of the same forgery types) than those seen during training, and
3) VideoSham which contains both new manipulation types and methods.

Tab.~\ref{table:results_on_joint_dataset} presents the results of our experiment. These results show that our approach achieves the strongest detection and localization performance on all categories.
In particular, on \pooleddsabrv-IND, we obtained the top detection and localization performance.
While a retrained version of MVSS-Net achieves the second best results in the in-distribution subset, its performance is substantially lower on the out-of-distribution subset (\pooleddsabrv-OOD).
By contrast, our approach both significantly outperforms others and maintains strong performance without suffering a significant drop when evaluated on new manipulation methods in \pooleddsabrv-OOD.
Our results on \pooleddsabrv-OOD show that \ournetworkname can generalize well to identify different versions of manipulations than those seen in training.
This trend also holds true for the VideoSham~\cite{VideoSham} dataset, whose forgeries are professionally made to be highly realistic using a diverse and unknown set of advanced techniques. 
%


\subsection{Results for Single-Manipulation Experiments}
\label{subsec:single_manip_eval}

Next, we conducted a set of experiments to evaluate \ournetworkname's ability to detect and localize a single targeted manipulation. We then compared its performance against detectors designed specifically for each type of manipulation. The results of these experiments show that \ournetworkname rivals competing specialized systems in most scenarios.

\input{tables/inpainting_results}

\subheader{Inpainting Forgeries Evaluation}
Our performance on AI-guided inpainting is shown in Tab.~\ref{table:inpainting_results}.
These results show that our approach achieves the best detection performance across all datasets. Similarly, we achieve the best localization performance on all testing sets with the exception of DAVIS-E2FGVI-HQ, where DVIL~\cite{DVIL} is slightly better (0.27 vs 0.21 F1).
We note that while existing systems use some form of temporal analysis, they overlooked both spatial evidence as well as temporal changes in forensic residuals~\cite{VIDNet, DVIL}. By contrast, \ournetworkname utilizes our new temporal residuals in combination with optical flow residuals and other spatial evidence, resulting in stronger performance.

\subheader{Splicing and Editing Forgeries Evaluation}
Tab.~\ref{table:splicing_editing_results} shows that our approach achieves the best detection mAP score on all testing sets. We also obtained the best localization F1 score on all datasets except for VCMS, where we are second place.
Notably, our performance remains consistently strong on both in-distribution and out-of-distribution datasets.
By contrast, the strongest competitor on the in-distribution datasets, a retrained MVSS-Net, performs significantly worse on the out-of-distribution datasets.
Since this network also exhibit this behavior in our experiment in Sec.~\ref{subsec:mult_manip_eval}, MVSS-Net likely overfits to the specific manipulation algorithms in training and has issues capturing generic evidence of different  forgery types.
Additionally, VideoFACT~\cite{VideoFACT} performed well on both in and out-of-distribution data because it is designed as a general video forensic network. However, unlike \ournetworkname, VideoFACT is unable to exploit any temporal information. This accounts for its lower performance when compared to \ournetworkname.

\input{tables/splicing_editing_results}

\subheader{Deepfake Forgeries Evaluation}
Tab.~\ref{table:deepfakes_results} shows the evaluation of our approach and others on detecting and localizing deepfaked videos.
When we evaluated on in-distribution datasets, we achieved the top performance on the DFD and F2F datasets, and remained competitive on the DF dataset.
When tested against new and unseen generation methods (NT, FS, FSH), our approach achieves the best performance on the NT dataset and second best performance on the FSH dataset. On the FS dataset, our performance is comparable with other state-of-the-art deepfake detectors like UCF~\cite{UCF}, SPSL~\cite{LiuCVPR2021}, SRM~\cite{LuoCVPR2021}, and SBI~\cite{SBI}.
Notably, while we did not train on NT or FS like other competing detectors, our approach is still able to obtain good performance on these datasets. Additionally, our network is general and \textit{can work on all video without needing a face detector to first locate a human face}. Even though face detectors are ubiquitous, they can still miss-detect, which adds as a source of error for existing deepfake detectors.

\input{tables/deepfakes_results}

%% file: tables/pooled_results.tex
\begin{table}[!t]
	\caption{Detection and localization performance on the \pooleddsname (\pooleddsabrv) and VideoSham~\cite{VideoSham} dataset. (R) denotes retrained.}
	\label{table:results_on_joint_dataset}
	\pullup
	\resizebox{1.0\linewidth}{!}{
		\begin{tblr}{
				width = 1.0\linewidth,
				colspec = {|m{16mm}|m{27mm}|m{7mm}|m{7mm}|m{7mm}|m{7mm}|m{7mm}|m{7mm}|},
				row{2} = {c},
				cell{1}{1} = {r=2}{},
				cell{1}{2} = {r=2}{},
				cell{1}{3} = {c=2}{0.30\linewidth,c},
				cell{1}{5} = {c=2}{0.32\linewidth,c},
				cell{1}{7} = {c=2}{0.30\linewidth,c},
				cell{-}{3-8} = {c},
				cell{3}{1} = {r=4}{},
				cell{7}{1} = {r=3}{},
				cell{10}{1} = {r=3}{},
				cell{13}{1} = {r=4}{},
				vline{1-3,5,7,9} = {1,2,3-15}{},
				hline{1} = {-}{},
				hline{2} = {-}{},
				hline{15,16} = {-}{},
				hline{3,7,10,13,17} = {-}{},
			}
			\textbf{Manip. Group}
			& \textbf{Method} & \textbf{\pooleddsabrv-IND} & & \textbf{\pooleddsabrv-OOD} & & \textbf{VideoSham} & \\
			&  & \textit{mAP} & \textit{F1} & \textit{mAP} & \textit{F1}& \textit{mAP} & \textit{F1} \\
			Splice/Edit
			& FSG~\cite{FSG} & 0.50 & 0.18 & 0.40 & 0.23 & 0.51 & 0.08 \\
			& MantraNet~\cite{ManTra-Net} & 0.51 & 0.16 & 0.46 & 0.08 & 0.49 & 0.07 \\
			& MVSS-Net~\cite{MVSS-Net} & 0.66 & 0.29 & 0.45 & 0.14 & 0.56 & 0.10 \\
			& MVSS-Net (R) & 0.94 & 0.77 & 0.63 & 0.09 & 0.52 & 0.01 \\
			Inpainting
			& VIDNet (R)~\cite{VIDNet} & 0.59 & 0.81 & 0.54 & 0.52 & 0.52 & 0.09 \\ 
			& DVIL~\cite{DVIL} & 0.58 & 0.25 & 0.53 & 0.22 & 0.53 & 0.07 \\ 
			& DVIL (R) & 0.50 & 0.57 & 0.54 & 0.50 & 0.49 & 0.03 \\ 
			Deepfake
			& SBI~\cite{SBI} & 0.66 & \ldash & 0.74 & \ldash &  0.54 & \ldash \\ 
			& MADD~\cite{MADD} & 0.57 & \ldash & 0.59 & \ldash &  0.56 & \ldash \\ 
			& CE.ViT~\cite{CE.ViT} & 0.58 & \ldash & 0.67 & \ldash &  0.55 & \ldash \\ 
			General
			& VideoFACT~\cite{VideoFACT} & 0.78 & 0.36 & 0.74 & 0.28 & 0.54 & 0.07 \\
			& VideoFACT (R) & 0.88 & 0.50 & 0.79 & 0.39 & 0.55 & 0.08 \\
			& Ensemble of (R)* & 0.90 & 0.54 & 0.73 & 0.44 & 0.52 & 0.11 \\
			& \textbf{Ours} & \best{0.95} & \best{0.77} & \best{0.91} & \best{0.59} & \best{0.63} & \best{0.14}
		\end{tblr}
	}
	\pulluppp\pullupp
\end{table}

%% file: tables/inpainting_results.tex
\begin{table}[!t]
	\pullup
    \caption{Detection and localization performance on inpainting datasets in \pooleddsabrv-IND and \pooleddsabrv-OOD. (R) denotes retrained.}
    \label{table:inpainting_results}
    \pullup
    \resizebox{1.0\linewidth}{!}{
        \begin{tblr}{
            width = \linewidth,
            colspec = {m{25mm}m{8mm}m{6mm}m{9mm}m{6mm}m{8mm}m{6mm}},
            row{2} = {c},
            cell{1}{1} = {r=3}{},
            cell{1}{2} = {c=2}{c},
            cell{1}{4} = {c=4}{c},
            cell{2}{2} = {c=2}{0.23\linewidth,c},
            cell{2}{4} = {c=2}{0.30\linewidth,c},
            cell{2}{6} = {c=2}{0.30\linewidth,c},
            cell{-}{2-7} = {c},
            vline{1-8} = {-}{},
            hline{1,3,4,9,10} = {-}{},
            hline{2} = {2-7}{},
        }
            \textbf{Method}
            & \textbf{\pooleddsabrv-IND} & & \textbf{\pooleddsabrv-OOD} & \\
            & \textbf{DEVIL} &
            & \textbf{DAVIS-E2FGVI-HQ} &
            & \textbf{DAVIS-FuseFormer} & \\
            & mAP & F1 & mAP & F1 & mAP & F1\\
            VIDNet (R)~\cite{VIDNet}    & 0.65 & 0.56 & 0.66 & 0.07 & 0.67 & 0.01 \\
            DVIL~\cite{DVIL}          	& 0.59 & 0.25 & 0.63 & \best{0.27} & 0.61 & 0.18 \\
            DVIL (R)                	& 0.58 & 0.35 & 0.53 & 0.10 & 0.54 & 0.10 \\
            VideoFACT~\cite{VideoFACT}  & 0.60 & 0.19 & 0.68 & 0.25 & 0.63 & 0.16 \\
            VideoFACT (R)           	& 0.76 & 0.39 & 0.69 & 0.08 & 0.78 & 0.06 \\
            \textbf{Ours}           	& \best{0.82} & \best{0.58} & \best{0.74} & 0.21 & \best{0.85} & \best{0.28}
        \end{tblr}
    }
	\pulluppp
\end{table}

%% file: tables/splicing_editing_results.tex
\begin{table}[!t]
	\pullup
    \caption{Detection and localization performance on splicing \& editing subsets in \pooleddsabrv-IND and \pooleddsabrv-OOD subset. (R) denotes the retrained version of the method.}
    \label{table:splicing_editing_results}
    \pullup
    \resizebox{1.0\linewidth}{!}{
    \begin{tblr}{
        width = \linewidth,
        colspec = {m{25mm}m{6.5mm}m{6.5mm}m{6.5mm}m{6.5mm}m{9mm}m{6.5mm}m{6.5mm}m{6.5mm}},
        row{2} = {c},
        cell{1}{1} = {r=3}{},
        cell{1}{2} = {c=4}{c},
        cell{1}{6} = {c=4}{c},
        cell{2}{2} = {c=2}{0.24\linewidth,c},
        cell{2}{4} = {c=2}{0.24\linewidth,c},
        cell{2}{6} = {c=2}{0.32\linewidth,c},
        cell{2}{8} = {c=2}{0.32\linewidth,c},
        cell{-}{2-9} = {c},
        vline{-} = {-}{},
        hline{1,3,4,10,11} = {-}{},
        hline{2} = {-}{},
    }
        \textbf{Method}
        & \textbf{\pooleddsabrv-IND} & & & & \textbf{\pooleddsabrv-OOD} & \\
        & \textbf{VCMS} &
        & \textbf{VPVM} &
        & \textbf{DAVIS Splice} &
        & \textbf{DAVIS Edit} & \\
        & mAP & F1 & mAP & F1 & mAP & F1 & mAP & F1 \\
        FSG~\cite{FSG}           	& 0.59 & 0.26 & 0.53 & 0.20 & 0.67 & 0.36 & 0.58 & 0.32 \\
        MantraNet~\cite{ManTra-Net} & 0.43 & 0.23 & 0.52 & 0.23 & 0.59 & 0.15 & 0.52 & 0.18 \\
        MVSS-Net~\cite{MVSS-Net}    & 0.89 & 0.60 & 0.69 & 0.38 & 0.84 & 0.49 & 0.56 & 0.17 \\
        MVSS-Net (R)            	& 0.99 & \best{0.87} & 0.98 & 0.86 & 0.66 & 0.13 & 0.62 & 0.07 \\
        VideoFACT~\cite{VideoFACT}  & 0.99 & 0.52 & 0.98 & 0.69 & 0.99 & 0.46 & 0.84 & 0.31 \\
        VideoFACT (R)           	& 0.99 & 0.49 & 0.96 & 0.65 & 0.99 & 0.45 & 0.86 & 0.32 \\
        \textbf{Ours}           	& \best{0.99} & 0.76 & \best{0.99} & \best{0.87} & \best{0.99} & \best{0.66} & \best{0.98} & \best{0.68}
    \end{tblr}
    }
	\pulluppp
\end{table}

%% file: tables/deepfakes_results.tex
\begin{table}[!t]
	\pullup
    \caption{\smallcaption Detection and localization performance on deepfake subsets in \pooleddsabrv-IND \& \pooleddsabrv-OOD. We provide additional benchmarks from DeepfakeBench~\cite{DeepfakeBench}. Each column's metrics are mAP/F1.}
    \label{table:deepfakes_results}
    \pullup
    \resizebox{1.0\linewidth}{!}{
        \begin{tblr}{
            width = \linewidth,
            colspec = {m{25mm}m{14.5mm}m{13.5mm}m{14.5mm}m{14.5mm}m{13.5mm}m{13.5mm}},
            row{2} = {c},
            cell{1}{1} = {r=2}{},
            cell{1}{2} = {c=3}{c},
            cell{1}{5} = {c=3}{c},
            vlines,
            hline{1,3} = {-}{},
            hline{2-3} = {2-7}{},
        }
            \textbf{Method}
            & \textbf{\pooleddsabrv-IND} & & & \textbf{\pooleddsabrv-OOD} & \\
            & \textbf{DFD}
            & \textbf{DF}
            & \textbf{F2F}
            & \textbf{NT}
            & \textbf{FS}
            & \textbf{FSH}\\
            SBI~\cite{SBI}       & 0.90/\ldash & 0.98/\ldash & 0.93/\ldash & \best{0.98}/\ldash & \best{0.99}/\ldash & \best{0.97}/\ldash \\
            MADD~\cite{MADD}      & 0.79/\ldash & 0.82/\ldash & 0.85/\ldash & 0.58/\ldash & 0.64/\ldash & 0.64/\ldash \\
            CE.ViT~\cite{CE.ViT}    & 0.82/\ldash & 0.92/\ldash & 0.79/\ldash & 0.74/\ldash & 0.92/\ldash & 0.83/\ldash \\
            VideoFACT~\cite{VideoFACT} & 0.54/0.07 & 0.53/0.23 & 0.55/0.15 & 0.69/0.24 & 0.56/0.09 & 0.80/0.45 \\
            VideoFACT (R)       & 0.75/0.36 & 0.89/0.68 & 0.84/0.71 & 0.92/0.65 & 0.68/0.66 & 0.61/0.50 \\
            \hline
            Xception~\cite{FF++}  & 0.82/\ldash & 0.98/\ldash & 0.98/\ldash & 0.94/\ldash & 0.98/\ldash & 0.62/\ldash \\
            X-ray~\cite{LiCVPR2020}     & 0.77/\ldash & 0.98/\ldash & \best{0.98}/\ldash & 0.93/\ldash & 0.98/\ldash & 0.65/\ldash \\
            UCF~\cite{UCF}       & 0.80/\ldash & \best{0.99}/\ldash & 0.98/\ldash & 0.94/\ldash & 0.98/\ldash & 0.65/\ldash \\
            SPSL~\cite{LiuCVPR2021}      & 0.81/\ldash & 0.98/\ldash & \best{0.98}/\ldash & 0.93/\ldash & 0.98/\ldash & 0.64/\ldash \\
            SRM~\cite{LuoCVPR2021}       & 0.81/\ldash & 0.97/\ldash & 0.97/\ldash & 0.93/\ldash & 0.97/\ldash & 0.60/\ldash \\
            \hline
            \textbf{Ours}       & \best{0.93/0.71} & 0.95/\best{0.74} & \best{0.98/0.77} & \best{0.95/0.82} & 0.92/\best{0.74} & 0.83/\best{0.72} \\
            \hline
        \end{tblr}
    }
	\pulluppp
\end{table}

%% file: sections/Discussion_v0.tex
\subsection{Effects of Video Compression}
\label{subsec:effects_of_crf}

We explore the effect of video compression on algorithms across 3 categories of video manipulations: Deepfake, Inpainting, and Splicing \& Editing. Fig.~\ref{fig:compression_comparison} shows the performance of top-performing algorithms on each manipulation type in the \pooleddsabrv-IND dataset under increasing compression strength.
The result shows that \ournetworkname consistently exhibits strong performance in the face of compression. For instance, \ournetworkname outperforms others across all compression levels on deepfake and inpainting, and shows much slower decline in performance on all manipulations.

\input{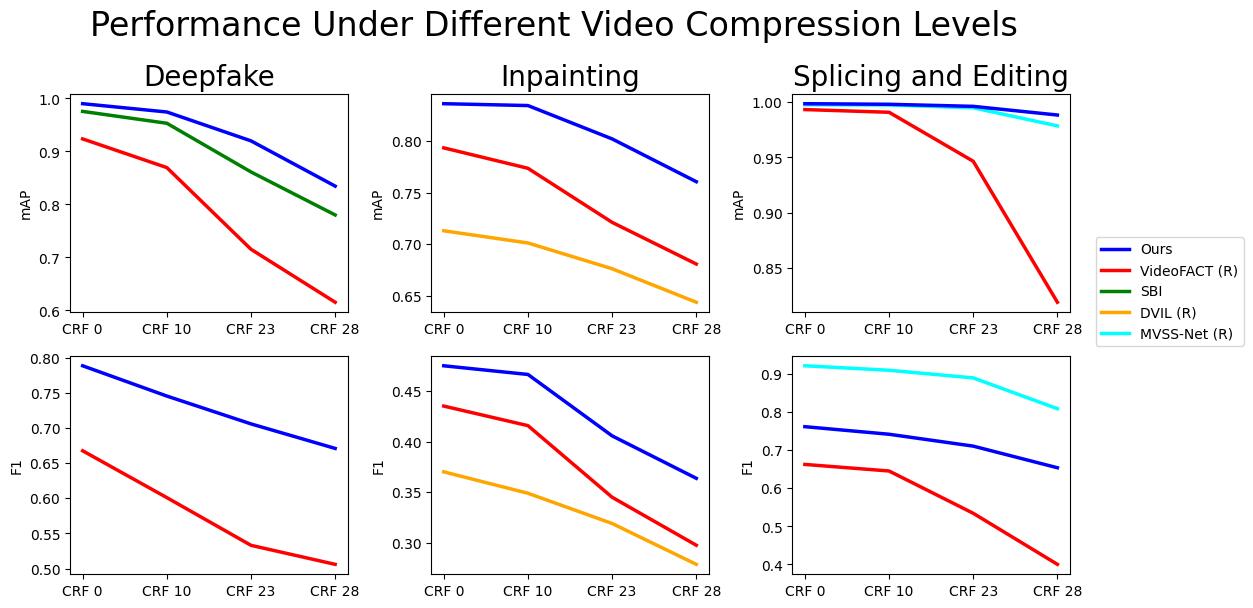}

\subsection{Failure Cases}
\label{subsec:failure_cases}

Our approach, while robust, faces limitations in certain conditions, such as: extreme video compression, advanced forgeries that are indistinguishable from real content, or new forgery artifacts not seen in training. Nevertheless, our network is designed to easily adapt through finetuning.

%% file: figures/crf_comparison.tex
\begin{figure}[!h]
	\pullupp
	\includegraphics[width=1.0\linewidth]{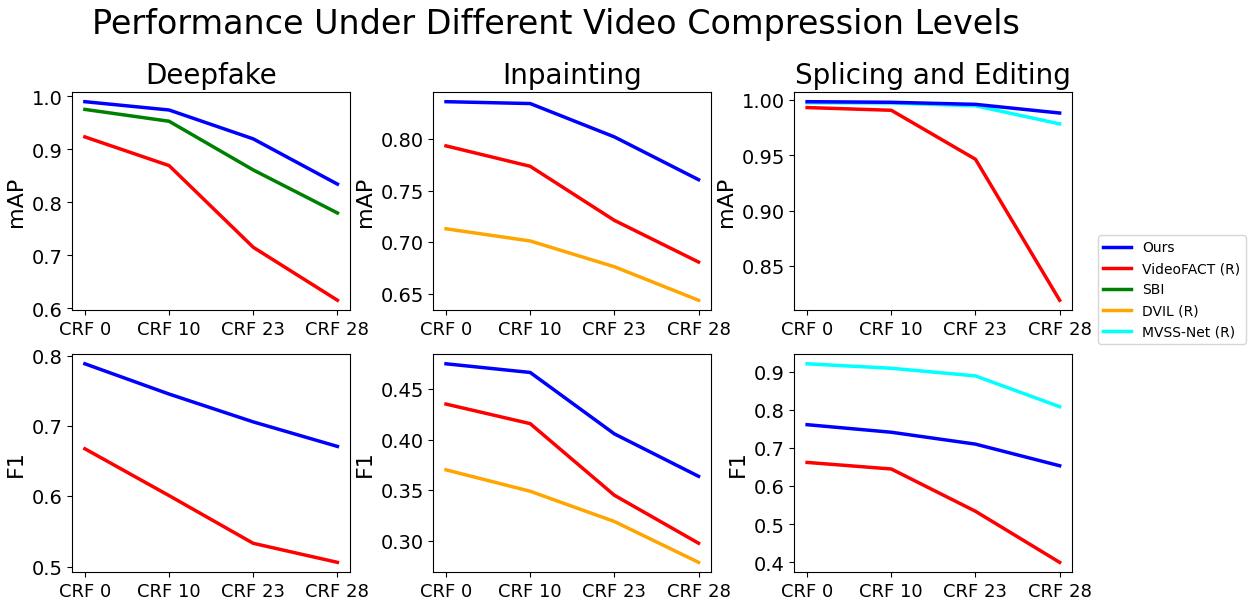}
	\caption{Effect of video compression on detection and localization performance in the \pooleddsabrv-IND dataset.}
	\label{fig:compression_comparison}
	\pulluppp
\end{figure}

%% file: sections/Ablation_v0.tex

\section{Ablation Study}
\label{sec:ablation}

We perform a comprehensive series of experiments to evaluate the effectiveness of each component in our proposed approach. The results are detailed in Tab.~\ref{table:ablation_evaluations}.

\subheader{Spatial Forensic Residuals}
We observe that our network performs much worse without the spatial forensic residual features. This is because spatial inconsistencies in forensic features play an important part in detection.

\input{tables/ablations}

\subheader{RGB Context Features}
We observe that removing the RGB context features decreases the performance of our network. This result demonstrates the importance of these features in contextualizing the forensic residual features.

\subheader{Temporal Forensic Residuals}
These features are designed to capture the change in temporal residual traces over time. Hence, removing them decreases the performance.

\subheader{Optical Flow Residuals}
Optical flow residual is important because it models the evolution of pixel values over time. Therefore, removing the Optical Flow module decreases the performance by 0.23 mAP and 0.43 F1.

\subheader{Standard Transformer}
As shown in Tab.~\ref{table:ablation_evaluations}, using a standard implementation of the Transformer not only significantly decreases performance, but also make our approach inflexible to input videos of different resolutions.

\subheader{M.S.H Transformer}
The M.S.H Transformer is needed for accurately modeling and detecting forensic inconsistencies across scales and modalities.
Tab.~\ref{table:ablation_evaluations} shows that without this module, our network's performance is much worse.

\subheader{Fine-to-Coarse}
Analyzing in a coarse-to-fine manner is crucial in preventing false-alarms from finer levels to propagate throughout our network. Hence, analyzing in a fine-to-coarse fashion reduced our network's performance.

%

%% file: tables/ablations.tex
\begin{table}[!t]
	\pullupp
    \caption{Ablation study of the components in our proposed network and their performance evaluations.}
    \label{table:ablation_evaluations}
    \pullup
    \centering
    \resizebox{0.75\linewidth}{!}{
        \begin{tblr}{
            width = \linewidth,
            colspec = {m{40mm}m{17mm}m{17mm}},
            row{2} = {c},
            cell{1}{1} = {r=2}{},
            cell{1}{2} = {c=2}{c},
            cell{3-11}{2-3} = {c},
            vline{1-4} = {-}{},
            hline{1,3-4,11} = {-}{},
            hline{2} = {2-3}{},
        }
            \textbf{Setup} & \textbf{\pooleddsabrv-IND} & \\
            & \textit{mAP} & \textit{F1}\\

            \textbf{Proposed}   		& \textbf{0.95} & \textbf{0.77} \\
            No Spatial Foren. Resid.   	& 0.64(-0.31) & 0.57(-0.20) \\
            No RGB Context 		        & 0.82(-0.13) & 0.52(-0.25) \\
            No Temporal Residual   		& 0.84(-0.11) & 0.59(-0.18) \\
            No Optical Flow Residual    & 0.72(-0.23) & 0.34(-0.43) \\
            Standard Transformer 		& 0.86(-0.09) & 0.56(-0.21) \\
            No M.S.H Transformer      	& 0.75(-0.20) & 0.52(-0.25) \\
            Fine-to-Coarse				& 0.90(-0.05) & 0.68(-0.09) 
        \end{tblr}
    }
	\pulluppp
\end{table}

%% file: sections/Conclusion_v0.tex

\section{Conclusion}
\label{sec:conclusion}

In this paper, we propose \ournetworkname, a versatile, multi-purpose network designed to address the complex challenge of detecting and localizing a broad set of video forgeries. Our approach integrates all evidence from multiple  forensic feature modalities, including new temporal forensic residual modality, and analyze them jointly. To do this, we leveraged a multi-scale hierarchical transformer accurately to identify forensic inconsistencies across different scales.
Our method outperforms specialized detectors in multi-manipulation contexts and performs better or equal to targeted systems in single-manipulation detection tasks.